\documentclass{article}

\usepackage[T1]{fontenc}
\usepackage[acronym,nowarn,section,nogroupskip,nonumberlist]{glossaries}

\glsdisablehyper{}
\newcommand{\acaps}[1]{{\scshape #1}}
\newacronym[\glslongpluralkey={Conditional Neural Processes}]{cnp}{\acaps{cnp}}{Conditional Neural Process}
\newacronym{mcmc}{\acaps{mcmc}}{Markov Chain Monte Carlo}
\newacronym{hmc}{\acaps{hmc}}{Hamiltonian Monte Carlo}
\newacronym{sgmcmc}{\acaps{sgmcmc}}{Stochastic Gradient Markov Chain Monte Carlo}
\newacronym{bnn}{\acaps{bnn}}{Bayesian Neural Network}
\newacronym[\glslongpluralkey={Deep Ensemble}]{de}{\acaps{de}}{Deep Ensembles}
\newacronym{l2e}{\acaps{l2e}}{Learning to Explore}

\newacronym{csgmcmc}{\acaps{csgmcmc}}{Cyclical Stochastic Gradient MCMC}
\newacronym{sghmc}{\acaps{sghmc}}{Stochastic Gradient Hamiltonian Monte Carlo}
\newacronym{sgld}{\acaps{sgld}}{Stochastic Gradient Langevin Dynamics}
\newacronym{psgld}{p\acaps{sgld}}{Preconditioned Stochastic Gradient Langevin Dynamics}
\newacronym{frn}{\acaps{frn}}{Filter Response Normalization}
\newacronym{sgrhmc}{\acaps{sgrhmc}}{Stochastic Gradient Relativistic Hamiltonian Monte Carlo}
\newacronym{sde}{\acaps{sde}}{Stochastic Differential Equation}
\newacronym{bma}{\acaps{bma}}{Bayesian Model Averaging}
\newacronym{mlp}{\acaps{mlp}}{Multi-Layer Perceptron}
\glsunset{mlp}
\newacronym[\glslongpluralkey={Evolution Strategies}]{es}{\acaps{es}}{Evolution Strategy}
\newacronym{psgmcmc}{\acaps{p-csgmcmc}}{Preconditioned Cyclic Stochastic Gradient MCMC}

\newacronym{tbptt}{\acaps{tbptt}}{Truncated BackPropagation Through Time}

\newacronym{ess}{\acaps{ess}}{Effective Sample Size}

\newacronym{msp}{\acaps{msp}}{Maximum Softmax Probability}

\newacronym{acc}{\acaps{acc}}{Accuracy}
\newacronym{nll}{\acaps{nll}}{Negative Log-Likelihood}
\newacronym{ece}{\acaps{ece}}{Expected Calibration Error}

\newacronym{kld}{\acaps{kld}}{Kullback–Leibler divergence}
\newacronym{ood}{\acaps{ood}}{Out-of-Distribution}
\usepackage{amsmath, amssymb, amsthm}
\usepackage{mathtools}
\usepackage{bbm}
\usepackage{dsfont}

\newcommand{\calD}{{\mathcal{D}}}

\newcommand{\calN}{{\mathcal{N}}}

\newcommand{\bbE}{\mathbb{E}}

\newcommand{\bbR}{\mathbb{R}}

\theoremstyle{plain}

\theoremstyle{definition}

\theoremstyle{remark}

\newcommand{\dee}{\mathrm{d}}

\DeclareMathOperator*{\argmax}{arg\,max}

\newcommand{\KL}{D_\mathrm{KL}}
\newcommand{\tr}{^\top}

\newcommand{\given}{\,|\,}

\def\[#1\]{\begin{equation}\begin{aligned}#1\end{aligned}\end{equation}}

\usepackage[usenames,dvipsnames]{xcolor}

\usepackage{graphicx}
\usepackage{caption}

\usepackage{booktabs, array}
\usepackage{makecell}
\usepackage{diagbox}
\newcolumntype{L}[1]{>{\raggedright\let\newline\\\arraybackslash\hspace{0pt}}m{#1}}
\newcolumntype{C}[1]{>{\centering\let\newline\\\arraybackslash\hspace{0pt}}m{#1}}
\newcolumntype{R}[1]{>{\raggedleft\let\newline\\\arraybackslash\hspace{0pt}}m{#1}}

\definecolor{citecolor}{RGB}{0,102,204}
\definecolor{linkcolor}{RGB}{190,105,30}
\definecolor{urlcolor}{RGB}{199,21,133}

\usepackage[colorlinks, linktoc=all,pagebackref=true]{hyperref}
\usepackage[all]{hypcap}
\hypersetup{citecolor=citecolor}
\hypersetup{linkcolor=linkcolor}
\hypersetup{urlcolor=urlcolor}
\usepackage[nameinlink,capitalise,noabbrev]{cleveref}
\creflabelformat{equation}{#2\textup{#1}#3}  
\crefname{section}{\S}{\S\S}

\usepackage[algoruled]{algorithm2e}
\usepackage{listings}
\usepackage{fancyvrb}
\fvset{fontsize=\small}

\usepackage{booktabs, array}
\usepackage{multirow}

\usepackage{duckuments}
\usepackage{graphicx,subcaption}

\usepackage[accepted]{icml2024}

\theoremstyle{plain}
\newtheorem{theorem}{Theorem}[section]

\theoremstyle{definition}

\newtheorem{assumption}[theorem]{Assumption}
\theoremstyle{remark}

\usepackage[textsize=tiny]{todonotes}

\icmltitlerunning{Learning to Explore for Stochastic Gradient MCMC}

\begin{document}

\twocolumn[
\icmltitle{Learning to Explore for Stochastic Gradient MCMC}



\icmlsetsymbol{equal}{*}

\begin{icmlauthorlist}
\icmlauthor{SeungHyun Kim}{equal,kaist}
\icmlauthor{Seohyeon Jung}{equal,comp}
\icmlauthor{Seonghyeon Kim}{kaist}
\icmlauthor{Juho Lee}{kaist,aitrics}
\end{icmlauthorlist}

\icmlaffiliation{kaist}{Kim Jaechul Graduate School of AI, Korea Advanced Institute of Science and Technology (KAIST), Daejeon, South Korea}
\icmlaffiliation{aitrics}{AITRICS, Seoul, South Korea}
\icmlaffiliation{comp}{Nexon Korea, Seongnam, South Korea(This work was conducted while the author was in KAIST.)}

\icmlcorrespondingauthor{SeungHyun Kim}{s.h.kim@kaist.ac.kr}
\icmlcorrespondingauthor{Seohyeon Jung}{heon2203@nexon.co.kr}
\icmlcorrespondingauthor{Juho Lee}{juholee@kaist.ac.kr}

\icmlkeywords{Machine Learning, ICML}

\vskip 0.3in
]



\printAffiliationsAndNotice{\icmlEqualContribution} 

\begin{abstract}

\glspl{bnn} with high-dimensional parameters pose a challenge for posterior inference due to the multi-modality of the posterior distributions. \gls{sgmcmc} with cyclical learning rate scheduling is a promising solution, but it requires a large number of sampling steps to explore high-dimensional multi-modal posteriors, making it computationally expensive. In this paper, we propose a meta-learning strategy to build \gls{sgmcmc} which can efficiently explore the multi-modal target distributions. Our algorithm allows the learned \gls{sgmcmc} to quickly explore the high-density region of the posterior landscape. Also, we show that this exploration property is transferrable to various tasks, even for the ones unseen during a meta-training stage. Using popular image classification benchmarks and a variety of downstream tasks, we demonstrate that our method significantly improves the sampling efficiency, achieving better performance than vanilla \gls{sgmcmc} without incurring significant computational overhead.

\end{abstract}

\glsresetall

\section{Introduction}
\label{main:sec:introduction}

Bayesian methods have received a lot of attention as powerful tools for improving the reliability of machine learning models. Bayesian methods are gaining prominence due to their ability to offer probability distributions over model parameters, thereby enabling the quantification of uncertainty in predictions. They find primary utility in safety-critical domains like autonomous driving, medical diagnosis, and finance, where the accurate modeling of prediction uncertainty often takes precedence over the predictions themselves. The integration of Bayesian modeling with (deep) neural networks, often referred to as \glspl{bnn}, introduces exciting prospects for the development of secure and trustworthy decision-making systems.

However, there are significant problems for the successful application of \glspl{bnn} in real-world scenarios. Bayesian inference in high-dimensional parameter space, especially for deep and large models employed for the applications mentioned above, is notoriously computationally expensive and often intractable due to the complexity of the posterior distribution. Moreover, posterior landscapes of \glspl{bnn} frequently display multi-modality, where multiple high density regions exist, posing a significant challenge to efficient exploration and sampling. Due to this difficulty, the methods that are reported to work well for relatively small models, for instance, variational inference~\citep{blei2017variational} or \gls{hmc}~\citep{neal2011mcmc}, can severly fail for deep neural networks trained with large amount of data, when applied without care. 

Recently, \gls{sgmcmc} methods~\citep{welling2011bayesian,chen2014stochastic,ma2015complete} have emerged as powerful tools for enhancing the scalability of approximate Bayesian inference. This advancement has opened up the possibilities of applying Bayesian methods to large-scale machine learning tasks. \gls{sgmcmc} offers a versatile array of methods for constructing Markov chains that converge towards the target posterior distributions. The simulation of these chains primarily relies on stochastic gradients, making them particularly suitable for \glspl{bnn} trained on large-scale datasets. However, despite the notable successes of \gls{sgmcmc} in some \gls{bnn} applications \citep{welling2011bayesian,chen2014stochastic,ma2015complete,zhang2020csgmcmc}, there remains a notable challenge. Achieving optimal performance often demands extensive engineering efforts and hyperparameter tuning. This fine-tuning process typically involves human trial and error or resource-intensive cross-validation procedures. Furthermore, it's worth noting that even with the use of \gls{sgmcmc} methods, there remains room for improvement in efficiently exploring multi-modal posterior distributions. As a result, in practical applications, a trade-off between precision and computational resources often becomes necessary.


To address these challenges, we introduce a novel meta-learning framework tailored to promote the efficient exploration  of \gls{sgmcmc} algorithms. Traditional \gls{sgmcmc} methods often rely on handcrafted design choices inspired by mathematical or physics principles. Recognizing the pivotal role these design components play in shaping the trade-off between exploration and exploitation within \gls{sgmcmc} chains, we argue in favor of learning them directly from data rather than manually specifying them. To achieve this, we construct neural networks to serve as meta-models responsible for approximating the gradients of kinetic energy terms. These meta-models are trained using a diverse set of \glspl{bnn} inference tasks, encompassing various datasets and architectural configurations. Our proposed approach, termed \gls{l2e}, exhibits several advantageous properties, including better mixing rates, improved prediction performance, and a reduced need for laborious hyperparameter tuning.

Our contributions can be summarized as follows:
\begin{itemize}
    \item We introduce \gls{l2e}, a novel meta-learning framework enhancing \gls{sgmcmc} methods. In contrast to conventional hand-designed approaches and meta-learning approach \citep{gong2018meta}, \gls{l2e} learns the kinetic energy term directly, offering a more data-driven and adaptable solution.
    \item   We present a multitask training pipeline equipped with a scalable gradient estimator for \gls{l2e}. This framework allows the meta-learned \gls{sgmcmc} techniques to generalize effectively across a wide range of tasks, extending their applicability beyond the scope of tasks encountered during meta-training.
    \item Using real-world image classification benchmarks, we demonstrate the remarkable performance of \glspl{bnn} inferred using the \gls{sgmcmc} algorithm discovered by \gls{l2e}, both in terms of prediction accuracy and sampling efficiency.
\end{itemize}

\section{Backgrounds}
\label{main:sec:backgrounds}

\subsection{\gls{sgmcmc} for Bayesian Neural Networks} 
\paragraph{Settings.}
In this paper, we focus on supervised learning problems with a training dataset $\calD = \{(x_i, y_i)_{i=1}^n$ with $x_i$ being observation and $y_i$ being label. Given a neural network with a parameter $\theta \in \bbR^d$, a likelihood $p(y\given x, \theta)$ and a prior $p(\theta)$ are set up, together defining an energy function $U(\theta) = - \sum_{i=1}^n\log p(y_i\given x_i, \theta) - \log p(\theta)$. The goal is to infer the posterior distribution $p(\theta\given\calD) \propto \exp(- U(\theta))$. When the size of the dataset $n$ is large, evaluating the energy function $U(\theta)$ or its gradient $\nabla_\theta U(\theta)$ may be undesirably costly as they require a pass through the entire dataset $\calD$. For such scenarios, \gls{sgmcmc}~\citep{welling2011bayesian,chen2014stochastic,ma2015complete} is a standard choice, where the gradients of the energy function $\nabla_\theta U(\theta)$ are approximated by a stochastic gradient computed from mini-batches. That is, given a mini-batch $B \subset \{1,\dots, n\}$ where $|B| \ll n$, an unbiased estimator of the full gradient $\nabla_\theta U(\theta)$ with $B$ can be computed as
\[
\nabla_\theta \tilde{U}(\theta) = -\frac{n}{|B|} \sum_{i \in B} \nabla_\theta \log p(y_i\given x_i,\theta) - \nabla_\theta \log p(\theta) \nonumber
\]

\paragraph{A complete recipe.} There may be several ways to build a Markov chain leading to the target posterior distribution. \citet{ma2015complete} presented a generic recipe that includes all the convergent \gls{sgmcmc} algorithms as special cases, constituting a complete framework. In this recipe, a parameter $\theta$ of interest is augmented with an auxiliary momentum variable $r$, and an \gls{sde} of the following form is defined for a joint variable $z = (\theta, r) \in \bbR^{2d}$ as follows.
\[\label{eq:sgmcmc_recipe}
&\dee z = \left[ -(D(z) + Q(z)) \nabla_z H(z) + \Gamma(z)\right] \dee t \\ &+ \sqrt{2 D(z)} \dee w_t \\
&H(z) := U(\theta) + g(\theta, r) \\  &\Gamma_i(z) := \sum_{j=1}^{2d} \frac{\partial}{\partial z_j}(D_{ij}(z) + Q_{ij}(z)),
\]
where $g(\theta, r)$ is the conditional energy function of the momentum $r$ such that $p(z) \propto \exp(-H(z))$ and $w_t$ is $2d$-dimensional Brownian motion. Here, $D(z) \in \bbR^{2d\times 2d}$ and $Q(z) \in \bbR^{2d\times 2d}$ are restricted to be positive semi-definite and skew-symmetric, respectively. Given this \gls{sde}, one can obtain a \gls{sgmcmc} algorithm by first substituting the full gradient $\nabla_z H(z)$ with a mini-batch gradient $\nabla_z \tilde H(z) = \nabla_z ( \tilde U(\theta) + g(\theta, r))$ and then discretizing it via a numerical solver such as sympletic Euler method. A notable example would be \gls{sghmc}~\citep{chen2014stochastic}, where $g(\theta, r) = \frac{1}{2}r\tr M^{-1} r$, $D(z) = \begin{bmatrix} 0 & 0 \\ 0 & C\end{bmatrix}$, and $Q(z) = \begin{bmatrix} 0 & -I \\ I & 0 \end{bmatrix}$ for some positive semi-definite matrices $M$ and $C$, leading to an algorithm when discretized with symplectic Euler method as follows.
\[
&r_{t+1} = r_t - \epsilon_t \nabla \tilde {U}(\theta_t) - \epsilon_t CM^{-1}r_t + \xi_t\\
& \theta_{t+1} = \theta_t + \epsilon_t M^{-1}r_{t+1},
\]
where $\xi_t \sim \calN(0, 2C\epsilon_t)$ and  $\epsilon_t$ is a step-size. 

The complete recipe includes interesting special cases that introduce adaptive preconditioners to improve the mixing of \gls{sgmcmc}~\citep{girolami2011riemann,li2016preconditioned,wenzel2020good}. For instance, \citet{li2016preconditioned} proposed \gls{psgld}, which includes RMSprop~\citep{RMSProp}-like preconditioning matrix in the updates:
\[
&\theta_{t+1} = \theta_t-\epsilon_t [ G(\theta_t)\nabla_\theta\tilde{U}(\theta) + \Gamma(\theta_t)] + \xi_t \\
& V(\theta_{t+1}) = \alpha V(\theta_t) + (1-\alpha) \frac{\nabla_\theta \tilde U(\theta_t)}{n} \odot \frac{\nabla_\theta \tilde U(\theta_t)}{n}\\
& G(\theta_{t+1}) = \mathrm{diag}(\mathbf{1} \oslash (\lambda \mathbf{1} + \sqrt{V(\theta_{t+1})}),
\]
where $\xi_t \sim \calN(0, 2G(\theta_t)\epsilon_t)$ and $\oslash, \odot$ denotes elementwise division and multiplication, respectively. \gls{psgld} exploits recent gradient information to adaptively adjust the scale of energy gradients and noise. However, this heuristical adjustment is still insufficient to efficiently explore the complex posteriors of \glspl{bnn}~\citep{zhang2020csgmcmc}.  


Recently, \citet{zhang2020csgmcmc} introduced cyclic learning rate schedule for efficient exploration of multi-modal distribution. The key idea is using the spike of learning rate induced by cyclic learning rate to escape from a single mode and move to other modes. However, in our experiment, we find that \gls{sgmcmc} with cyclical learning rate does not necessarily capture multi-modality and it also requires a large amount of update steps to move to other modes in practice. 


\paragraph{Prediction via Bayesian model averaging.} 
After inferring the posterior $p(\theta\given\calD)$, for a test input $x_*$, the posterior predictive is computed as
\[
p(y_*\given x_*, \calD) = \int_{\bbR^d} p(y_*\given x_*, \theta) p(\theta\given \calD) \dee \theta,
\]
which is also referred to as \gls{bma}. In our setting, having collected from the posterior samples $\theta_1,\dots, \theta_K$ from a convergent chain simulated from \gls{sgmcmc} procedure, the predictive distribution is approximated with a Monte-Carlo estimater,
\[
p(y_*\given x_*, \calD) \approx \frac{1}{K}\sum_{k=1}^K p(y_*\given x_*, \theta_k).
\]
As one can easily guess, the quality of this approximation depends heavily on the quality of the samples drawn from \gls{mcmc} procedure. For over-parameterzied deep neural networks that we are interested in, the target posterior $p(\theta\given\calD)$ is typically highly multi-modal, so simple \gls{sgmcmc} methods suffer from poor mixing; that is, the posterior samples collected from those methods are not widely spread throughout the parameter space, so it takes exponentially many samples to achieve desired level of accuracy for the approximation. Hence, a good \gls{sgmcmc} algorithm should be equipped with the ability to efficiently explore the parameter space, while still be able to stay sufficiently long in high-density regions. That is, it should have a right balance between exploration-exploitation.

\paragraph{Meta Learning}Meta-learning, or learning to learn, refers to the algorithm that learns the useful general knowledge from source tasks that can transfer to the unseen tasks. Most meta-learning algorithms involves two levels of learning: an inner-loop and outer-loop~\citep{metz2018meta}. Inner-loop usually contains the training procedure of particular task. In our work, inner-loop for our meta-training is iteratively update the model parameter $\theta$ by running \gls{sgmcmc} with learnable transition kernel. Outer-loop refers to the training procedure of meta-parameter $\phi$, which is done by minimizing meta-objective $L(\phi)$. 


\paragraph{Meta Learning and MCMC}

There exists line of work that parameterize the transition kernel of MCMC with trainable function for various purposes. \citet{levy2017generalizing} used learnable invertible operator to automatically design the transition kernel of HMC for good mixing. For \gls{sgmcmc}, \citet{gong2018meta} parameterized the curl matrix and acceleration matrix using neural networks under the framework of \citet{ma2015complete}. Although \citet{gong2018meta} is the closest work for our method, this work did not verified its task generalization, i.e., evaluated only on the dataset that was used for training. Moreover, the scale of the experiments in \citet{gong2018meta} are limited to small-sized network architectures. We further demonstrate the limitation of \citet{gong2018meta} in simulating large scale multi-modal \glspl{bnn} posterior, along with detailed discussions in \cref{app:sec:metasgmcmc}. To the best of our knowledge, our work firstly proposes the method to meta-learn \gls{sgmcmc} that can generalize to unseen datasets and scale to large-scale \glspl{bnn}.

\paragraph{Limitation of Meta-\gls{sgmcmc} \citep{gong2018meta}}
Meta-\gls{sgmcmc} aims to learn the diffusion matrix $D(z)$ and the curl matrix $Q(z)$ in \cref{eq:sgmcmc_recipe} to build a \gls{sgmcmc} algorithm that can quickly converge to a target distribution. For this purpose, neural network $D_f(z)$ and $Q_f(z)$ are employed to model $D(z)$ and $Q(z)$. We point out that this parameterization has notable limitations, especially in terms of efficient exploration of high-dimensional multi-modal target distribution.

\begin{itemize}
    \item According to the recipe in \cref{eq:sgmcmc_recipe}, the diffusion and the curl that are dependent on $z$ involves the additional correction term $\Gamma(z)$. This hurts the scalability of the algorithm, as computing $\Gamma(z)$ involves computing the gradient of $D_f(z)$ and $Q_f(z)$ with respect to $z$. This amounts to a significant computational burden as the dimension of $z$ increases and involves finite-difference approximation. While Meta-\gls{sgmcmc} attempted to address this issue of computational cost through finite difference, but it still introduces additional computation with time complexity $O(HD)$ where $H$ is number of hidden units of neural network in meta-sampler and $D$ is dimension of $\theta$. Also, it is important to note that adding such approximation can negatively impact the convergence of the sampler.
    \item The objective function of Meta-\gls{sgmcmc} is the KL divergence between the target distribution and the distribution of the samples obtained from the Meta-\gls{sgmcmc} chain. This objective is not computed in closed-from and nor its gradient. \citet{gong2018meta} proposes to use Stein gradient estimator~\cite{li2018gradient} which is generally hard to tune. Moreover, \citet{gong2018meta} uses \gls{tbptt} for the gradient estimation, which is a biased estimator of the true gradient, and limited in the length of the inner step for meta-learning due to the memory consumption.
\end{itemize}

\section{Main contribution: Learning to Explore}
\label{main:sec:method}

\subsection{Overcome the limitations of Meta-\gls{sgmcmc}}

To avoid aforementioned limitations of Meta-\gls{sgmcmc}, we newly design the meta-learning method for \gls{sgmcmc} as follows:
\begin{itemize}
    \item

    We parameterize the gradients of the kinetic energy function and keep $D_f(z)$ and $Q_f(z)$ independent of $z$. We can avoid the additional computational cost of computing $\Gamma(z)$ without sacrificing the flexibility of the sampler.
    \item 
    We use a new meta-objective called \gls{bma} meta-loss which is the Monte-Carlo estimator of the predictive distribution to enhance the exploration and performance of the learned sampler. Both of our meta-objective and its gradient can be computed in closed-form. Also, we employ \gls{es}~\citep{salimans2017evolution,metz2019understanding} for gradient estimation, which is an unbiased estimator of the true meta-gradient. This also consumes significantly less memory compared to analytic gradient methods, allowing us to keep the length of the inner-loop much longer during meta-learning.

\end{itemize}

\subsection{Meta-learning framework for \gls{sgmcmc}}
Instead of using a hand-designed recipe for \gls{sgmcmc}, we aim to \emph{learn} the proper \gls{sgmcmc} update steps through meta learning. The existing works, both the methods using hand-designed choices or meta-learning~\citep{gong2018meta}, try to determine the forms of the matrices $D(z)$ and $Q(z)$ while keeping the kinetic energy $g(\theta, r)$ as simple Gaussian energy function, that is, $g(\theta, r) = r\tr M^{-1}r / 2$. This choice indeed is theoretically grounded, which can be shown to be optimal when the target distribution is Gaussian~\citep{betancourt2017conceptual}, but may not be optimal for the complex multi-modal posteriors of \glspl{bnn}.
We instead choose to learn $g(\theta, r)$ while keeping $D(z)$ and $Q(z)$ as simple as possible. We argue that the meta-learning approach based on this alternative parameterization is more effective in learning versatile \gls{sgmcmc} procedure that scales to large \glspl{bnn}. 

More specifically, we parameterize the gradients of the kinetic energy function $\nabla_\theta g(\theta, r)$ and $\nabla_r g(\theta, r)$ with neural networks $\alpha_\phi(\theta, r)$ and $\beta_\phi(\theta, r)$ respectively, and set $D(z)$ and $Q(z)$ as in \gls{sghmc}. The update step of \gls{sgmcmc}, when discretized with symplectic Euler method is,
\[
& r_{t+1} = r_t - \epsilon_t [ \nabla_\theta \tilde{U}(\theta_t) + \alpha_\phi(\theta_t, r_t) + C\beta_\phi(\theta_t, r_t)] + \xi_t \\
& \theta_{t+1} = \theta_t + \epsilon_t \beta_\phi(\theta_t, r_{t+1}). \label{eq:update}
\]


where $\xi_t \sim \calN(0, 2C\epsilon_t)$. Since we parameterize $\nabla g(\theta,r)$ and do not explicitly define the form of $g(\theta,r)$, we make the following assumptions about the underlying function $g(\theta,r)$.

\begin{assumption}\label{as:1}
  There exists an energy function $g(\theta,r)$ whose gradients with respect to $\theta,r$ are $\alpha_\phi(\theta, r)$ and $\beta_\phi(\theta, r)$ respectively, and $\int \exp(-g(\theta,r))dr = \text{constant}.$
\end{assumption}


We also present another version of \gls{l2e} which does not require additional assumptions in \cref{app:sec:kineticl2e}. The neural networks $\alpha_\phi$ and $\beta_\phi$ are parameterized as two-layer \glspl{mlp} with 32 hidden units. Specifically,  $\alpha_\phi$ and $\beta_\phi$ are applied to each dimension of parameter and momentum independently, similar to the commonly used learned optimizers~\citep{andrychowicz2016learning,metz2019understanding}. Again, following the common literature in learned optimizers~\citep{metz2019understanding}, for each dimension of the parameter and momentum, we feed the corresponding parameter and momentum values, the stochastic gradients of energy functions for that element, and running average of the gradient at various time scales, as they are reported to encode the sufficient information about the loss surface geometry. See \cref{app:sec:meta} for implementation details of $\alpha_\phi$ and $\beta_\phi$. By leveraging this information, we expect our meta-learned \gls{sgmcmc} procedure to capture the multi-modal structures of the target posteriors of \glspl{bnn}, and thus yielding a better mixing method.

\subsection{Meta-Objective and Optimization}
\paragraph{Objective functions for meta-learning.}
Meta-objective should reflect the meta-knowledge one wants to learn. We design the meta-objective based on the hope that samples collected through \gls{sgmcmc} should be good at approximating the posterior predictive $p(y_*|x_*,\calD)$. In order to achieve this goal, we propose the meta-objective called \gls{bma} meta-loss. After the sufficient number of inner-updates, we collect $K$ parameter samples with some interval between them (thinning). Let $\theta_k(\phi)$ be the $k^{th}$ collected parameter, and we compute the Monte-Carlo estimator of the predictive distribution and use it as a meta-objective function (note the dependency of $\theta_k$ on the meta-parameter $\phi$, as it is a consequence of learning \gls{sgmcmc} with the meta-parameter $\phi$).
\[
L(\phi) = - \log \frac{1}{K} \sum_{k=1}^K p(y_*\given x_*, \theta_k(\phi)),
\]
where $(x_*, y_*)$ is a validation data point.




\begin{algorithm}[t]
    \begin{algorithmic}[1]
        \caption{Meta training procedure}\label{alg:outer}
        \STATE {\bfseries Input:} Task distribution $P(\mathcal{T})$, inner iterations $N_\text{inner}$, outer iterations $N_\text{outer}$, step size $\epsilon$, noise scale $\sigma^2$, initial meta-parameter $\phi_0$.
        \STATE {\bfseries Output:} Meta parameter $\phi$.
        \FOR{$j=1,\dots, N_{\normalfont\text{outer}}$}
        \STATE Sample task $T_{i} \sim P(\mathcal{T})$
        \STATE Initialize model parameter $\theta_0$ for $T_{i}$
        \STATE Sample $\eta \sim \calN(0,\sigma^2 I)$ 
        \STATE $L(\phi + \eta) \gets \text{InnerLoop}(\theta_0, \phi + \eta, \epsilon , N_\text{inner})$ 
        \STATE $L(\phi - \eta)  \gets \text{InnerLoop}(\theta_0, \phi - \eta, \epsilon,N_\text{inner})$
        \STATE $\nabla_{\phi}L \gets \frac{1}{2\,\sigma^2}\eta\, (L(\phi +\eta) - L(\phi - \eta))$
        \STATE $\phi \gets \phi - \gamma \nabla_{\phi}L(\phi)$
        \ENDFOR
    \end{algorithmic}
    
\end{algorithm}


\paragraph{Gradient estimation for meta-objective.}
Estimating the meta-gradient $\nabla_\phi L(\phi)$ is highly non-trivial~\citep{metz2018meta,metz2019understanding}, especially when the number of inner update steps is large. For instance, a na\"ive method such as backpropagation through time would require memory grows linearly with the number of inner-steps, so become easily infeasible for even moderate sized models. One might consider using the truncation approximation, but that would result in a biased gradient estimator. Instead, we adapt \gls{es}~\citep{salimans2017evolution} with antithetic sampling scheme, which has been widely used in recent literature of training learned optimizer. \citet{metz2019understanding} showed that unrolled optimization with many inner-steps can lead to chaotic meta-loss surface and \gls{es} is capable of relieving this pathology by employing smoothed loss,  
\[
L(\phi) = \mathbb{E}_{\tilde{\phi} \sim \calN (\phi,\sigma^2 I)} \left[ L(\tilde \phi) \label{eq:smloss} \right]
\] 
where $\sigma^2$ determines the degree of smoothing. Also, antithetic sampling is usually applied to reduce the estimation variance of $\nabla_{\phi} L(\phi)$. Through log-derivative trick, we can get unbiased estimator of \eqref{eq:smloss}, \[\hat{g} = \frac{1}{N} \sum_{i=1}^N L(\phi+\eta_i)\frac{\eta_i}{\sigma^2} \] where $\eta_i \overset{\mathrm{i.i.d}}{\sim} \calN(0,\sigma^2I)$.  In addition, we can get another unbiased estimator $\hat{g}^{-1} =-\frac{1}{N} \sum_{i=1}^N L(\phi - \eta_i)\frac{\eta_i}{\sigma^2} $ by reusing the negative of $\eta_i$. By taking the average of two estimators, we can obtain the following gradient estimator. 
\[
\hat{\nabla}_{\phi}L(\phi) = \frac{1}{N}\sum_{i=1}^N \left[ \frac{L(\phi+\eta_i) - L(\phi -\eta_i)}{2\sigma^2} \right] \eta_i
\] 
The estimator is also amenable to parallelization, improving the efficiency of gradient computation.

\subsection{Meta training procedure}
\begin{figure*}[ht]
    \centering
    \begin{subfigure}[b]{0.44\textwidth}
        \centering
        \includegraphics[width=\textwidth]{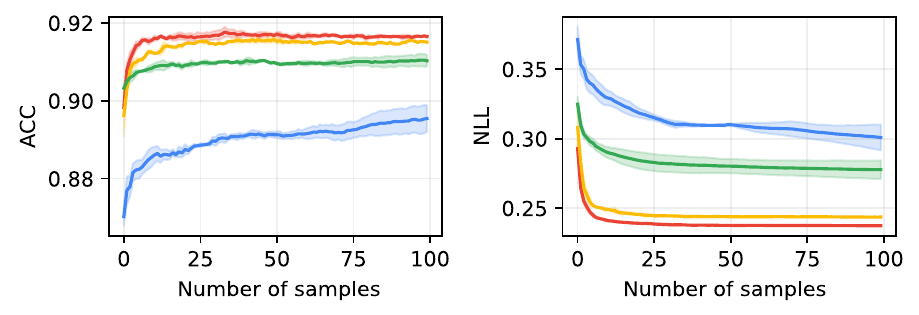}
        \caption{fashion-MNIST (seen task)}    
        \label{fig:main:fmnist}
    \end{subfigure}
    \hfill
    \begin{subfigure}[b]{0.55\textwidth}  
        \centering 
        \includegraphics[width=\textwidth]{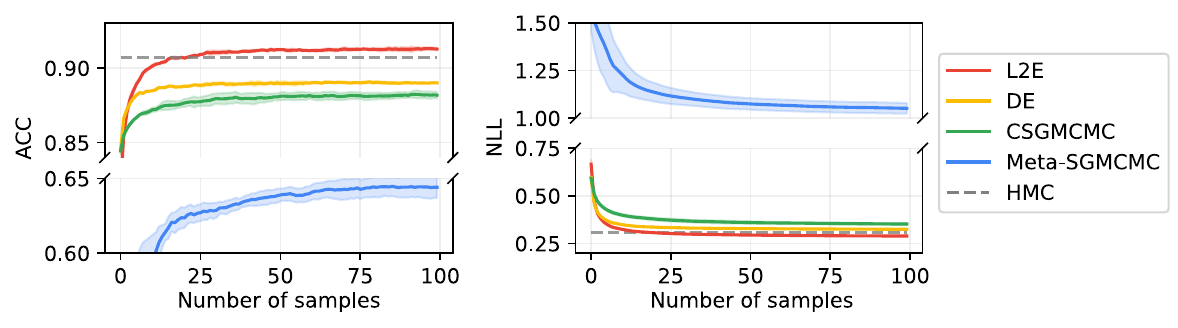}
        \caption{CIFAR-10}   
        \label{fig:main:c10}
    \end{subfigure}
    \begin{subfigure}[b]{0.44\textwidth}   
        \centering 
        \includegraphics[width=\textwidth]{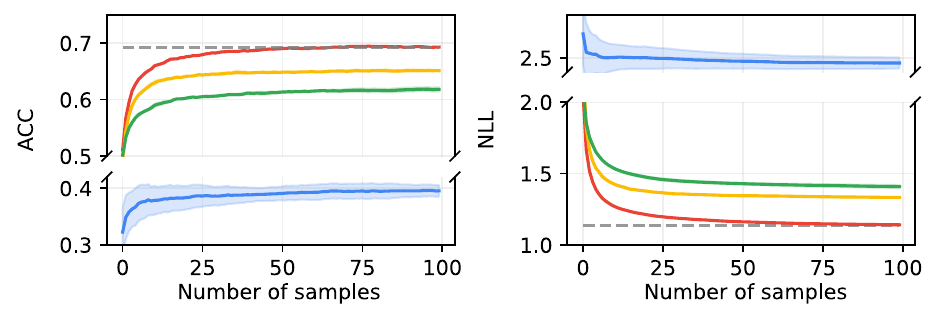}
        \caption{CIFAR-100} 
        \label{fig:main:c100}
    \end{subfigure}
    \hfill
    \begin{subfigure}[b]{0.55\textwidth}   
        \centering 
        \includegraphics[width=\textwidth]{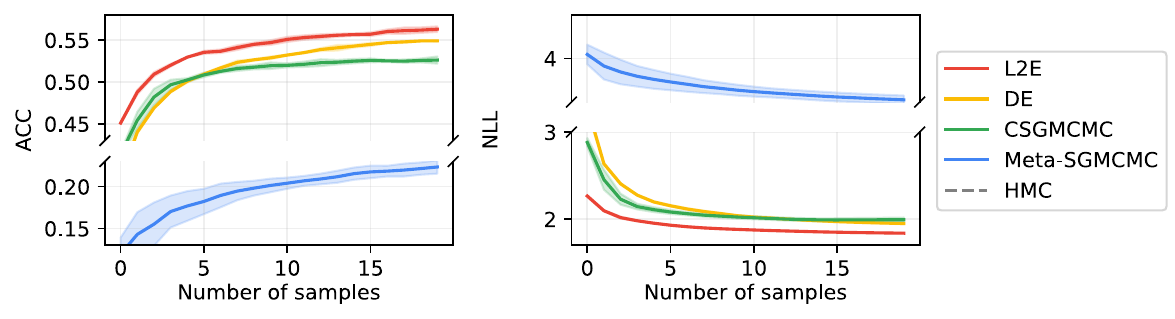}
        \caption{Tiny-ImageNet} 
        \label{fig:main:tiny}
    \end{subfigure}
    \caption{Predictive performance trend of each method as the number of samples for BMA increases. L2E exhibits superior predictive accuracy compared to other baseline methods. Note that only the fashion-MNIST dataset is included in meta-training task distribution. We also plot the performance of \gls{hmc} reference samples from \citet{izmailov2021bayesian} as dashed line for CIFAR-10 and CIFAR-100. For meta-training details of Meta-\gls{sgmcmc}, please refer to \cref{app:sec:metasgmcmctraining}.} 
    \label{fig:main}
\end{figure*}

\paragraph{Generic pipeline.}
General process of meta-training is as follows. First, for each inner-loop, we sample a task from the pre-determined task distribution. An inner-loop starts from an randomly initialized parameter and iteratively apply update step \eqref{eq:update} to run a single chain of \gls{sgmcmc}. Please refer to \cref{alg:inner} for detailed description. In the initial stage of meta-training, the chains from these inner loops show poor convergence, but the performance improves as training progresses. Similar to general Bayesian inference, we consider the early part of the inner loop as a burn-in period and collect samples from the end of the inner-loop at regular intervals when evaluating the meta-objective. This training process naturally integrates the meta-learning and Bayesian inference in that mimicking the actual inference procedure of Bayesian methods in realistic supervised learning tasks. In \cref{fig:main} we show that \gls{l2e} achieve desired level of accuracy for the approximation of posterior predictive with relatively small number of samples. This result indicates that \gls{l2e} has successfully acquired the desired properties through meta-training.  

\paragraph{Multitask training for better generalization.}
In meta-learning, diversifying the task distribution is commonly known to enhance generalization. We include various neural network architectures and datasets in the task distribution to ensure that \gls{l2e} has sufficient generalization capacity. Also, we evaluate how the task distribution diversity affects the performance of \gls{l2e} in \cref{app:tab:ablation}.

\section{Experiments}
\label{main:sec:experiments}


In this section, we will evaluate the performance of \gls{l2e} in various aspects. Through extensive experiments, we would like to demonstrate followings:

\begin{itemize}
    \item \gls{l2e} shows excellent performance on real-world image classification tasks and seamlessly generalizes to the tasks not seen during meta-training.

    \item \gls{l2e} produces similar predictive distribution to \gls{hmc} as well as good mixing of \glspl{bnn} posterior distribution in weight space.  
    
    \item \gls{l2e} can effectively explore and sample from \glspl{bnn} posterior distribution, collecting diverse set of parameters both in weight space and function space.
\end{itemize}

\paragraph{Experimental details}
In this paragraph, we explain our experimental settings. We evaluate \gls{l2e} and baseline methods on 4 datasets: fashion-MNIST~\citep{xiao2017fashion}, CIFAR-10, CIFAR-100~\citep{krizhevsky2009learning} and Tiny-ImageNet~\citep{le2015tiny}. Please refer to \cref{app:sec:baseline} for full experimental setup and details. We report the mean and standard deviation of results over three different trials. Code is available at \url{https://github.com/ciao-seohyeon/l2e}.

\paragraph{Comparing \gls{l2e} with \gls{hmc}}
\gls{hmc} is popular \gls{mcmc} method since it can efficiently explore the target distribution and asymptotically converge to the target distribution, making it useful for inference in high-dimensional spaces. Due to its asymptotic convergence property, \gls{hmc} often considered as golden standard in Bayesian Inference. However, its computational burden harms the applicability of \gls{hmc} on modern machine learning tasks.
Recently, \citet{izmailov2021bayesian} ran \gls{hmc} on CIFAR-10 and CIFAR-100 datasets and released logits of \gls{hmc} as reference. For deeper investigation of mixing of \gls{l2e}, we compare \gls{l2e} with \gls{hmc}. 

In order to compare the results with reference HMC samples from \citet{izmailov2021bayesian}, we replicate the experimental setup of \citet{izmailov2021bayesian} including neural network architecture and dataset processing. The distinctive aspects of this setup are that they excluded data augmentation and replaced batch normalization layer in ResNet architecture to \gls{frn}~\citep{singh2020filter} to remove stochasticity in evaluating posterior distribution. We use the ResNet20-FRN architecture with Swish activations~\citep{ramachandran2017searching} and we use 40960 samples for training all methods. Also, we compare \gls{l2e} to \gls{hmc} 1-D synthetic regression experiments in \cref{app:sec:1d}. We also provide experimental results with data augmentation in \cref{app:sec:aug} with other discussions.


\begin{figure*}[ht]
\centering
\begin{subfigure}{0.3\textwidth}
    \includegraphics[width=\textwidth]{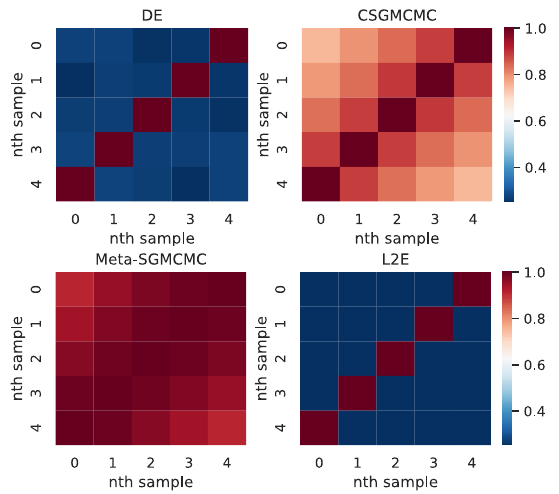}
    \caption{Cosine similarity between weights.}
    \label{fig:exp:cosine}
\end{subfigure}
\hfill
\begin{subfigure}{0.335\textwidth}
    \includegraphics[width=\textwidth]{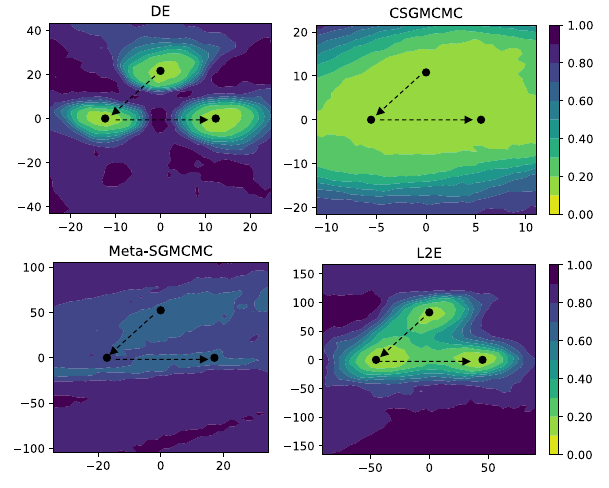}
    \caption{Test error ($\%$) around the samples.}
    \label{fig:exp:surface}
\end{subfigure}
\hfill
\begin{subfigure}{0.287\textwidth}
    \includegraphics[width=\textwidth]{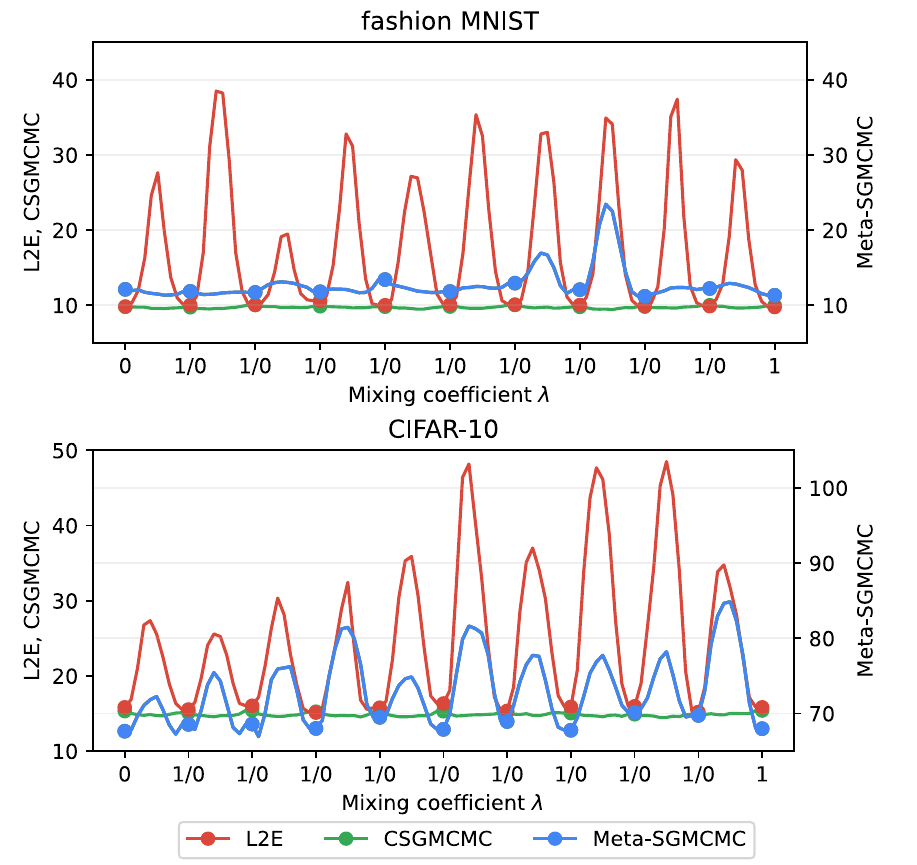}
    \caption{Test error ($\%$) along linear path}
    \label{fig:exp:tsne}
\end{subfigure}
\caption{Figures show multi-modality of various methods \gls{de}, \gls{csgmcmc}, Meta-\gls{sgmcmc} and \gls{l2e} with ResNet20-FRN on CIFAR-10. (a) shows cosine similarity between weights. (b) is loss surface as a function of model parameters in a 2-dimensional subspace spanned by solutions of each method. Colors represent the level of test error. (c) shows test error ($\%$) along linear path between a pair of parameters. Due to the inferior performance of Meta-\gls{sgmcmc}, the offset of test error adjusted individually in CIFAR-10.}
\label{fig:exp:multi_modality}
\end{figure*}

\paragraph{Task distribution for meta-training}
We construct set of meta-training tasks using various datasets and model architectures. We use these tasks to meta-train meta-\gls{sgmcmc} and \gls{l2e}. Specifically, we use MNIST, fashion-MNIST, EMNIST \citep{cohen2017emnist} and MedMNIST \citep{yang2021medmnist} as meta-training datasets. For model architecture, we fix the general structure with several convolution layers followed by readout MLP layer. For each outer training iteration, we randomly choose dataset and sample the configuration of architecture including number of channels, depth of the convolution layers and whether to use skip connections. See \cref{app:sec:meta} for detailed configuration of task distribution. For evaluation, we use same meta-parameter of \gls{l2e} and meta-\gls{sgmcmc} for all experiments to check the generalization ability of each method.

\subsection{Image classification results}
\cref{fig:main} illustrates the trend of predictive performance of each method as the number of samples for \gls{bma} increases. We confirm that \gls{l2e} outperforms other baselines in terms of predictive accuracy on completely unseen datasets and architectures during meta-training. Among datasets for evaluation, \emph{fashion-MNIST is the only dataset included in our task distribution.} Despite not having seen other datasets during meta-training, \gls{l2e} consistently outperforms other tuned baseline methods, showing that \gls{l2e} can scale and generalize well to unseen larger tasks. Specifically, only \gls{de} shows comparable predictive accuracy compared to \gls{l2e} in fashion-MNIST dataset. In general, \gls{l2e} shows rapid performance gain and outperforms other methods with relatively small number of samples. According to \cref{fig:exp:multi_modality}, \gls{l2e} collects diverse weights and functions and this diversity presumably affect the huge performance gain from \gls{bma}. 

On the other hand, Meta-\gls{sgmcmc} exhibits poor predictive accuracy over all experiments. It converges very slowly compared to other methods and stucks in the low density region. This tendency gets worse when the evaluation task is significantly different from the distribution of meta-training tasks. In fashion-MNIST task, the performance of Meta-\gls{sgmcmc} is still worse but relatively close to other methods than other tasks. Since learned sampler has seen fashion-MNIST dataset during meta-training, it adapts to this task to a moderate extent. However, it cannot produce even reasonable performance on other large scale tasks. We also evaluate \gls{l2e} in the same experimental setup of \citet{gong2018meta} and compare to the reported performance of Meta-\gls{sgmcmc} in \cref{app:sec:metasgmcmc}. Without having seen CIFAR-10 dataset during meta-training, \gls{l2e} significantly outperforms Meta-\gls{sgmcmc} also in the experimental setup of \citet{gong2018meta}.

\newsavebox\CBox 
\newcommand{\spm}[1]{\scriptstyle{\pm#1}}
\def\textBF#1{\sbox\CBox{#1}\resizebox{\wd\CBox}{\ht\CBox}{\textbf{#1}}}

\begin{table}[t]
    \centering
     \caption{Fidelity to predictive distribution given by \gls{hmc} samples from \citet{izmailov2021bayesian}.}
     \label{exp:tab:agreement}
     \vskip 0.15in
        \resizebox{\columnwidth}{!}{
        \begin{tabular}{llrrr}
     \toprule
     Metric & Dataset &  \gls{de}  & \gls{csgmcmc} & \gls{l2e} 
     \\
     \midrule
    \multirow{2}{*}{Agreement}
    & CIFAR-10 & 0.920$\spm{0.001}$ & 0.910$\spm{0.007}$ & \textbf{0.946}$\spm{0.002}$ \\
    & CIFAR-100 & 0.751$\spm{0.001}$ & 0.726$\spm{0.007}$ & \textbf{0.771}$\spm{0.003}$ \\
    \midrule
    \multirow{2}{*}{Total Var}
    & CIFAR-10 & 0.0103$\spm{0.0001}$ & 0.0108$\spm{0.0005}$ & \textbf{0.0075}$\spm{0.0001}$ \\
    & CIFAR-100 & 0.0026$\spm{0.0000}$ & 0.0024$\spm{0.0001}$ & \textbf{0.0022}$\spm{0.0000}$ \\
     \bottomrule
        \end{tabular}}
    \vskip -0.2in
\end{table}

\subsection{Similarity to the \gls{hmc} samples}

As mentioned above, we compare \gls{l2e} to \gls{hmc} samples from \citet{izmailov2021bayesian} to objectively evaluate and investigate whether \gls{l2e} has converged to target distribution. We measure the agreement between predictive distribution of \gls{hmc} and \gls{l2e}, and the difference of probability vector of two methods following \citet{izmailov2021bayesian}. Please refer to \cref{app:sec:experiments} for details of metrics. In \cref{exp:tab:agreement}, we confirm that \gls{l2e} makes higher fidelity approximation to predictive distribution of \gls{hmc} than other baselines both in CIFAR-10 and CIFAR-100. This implies that L2E shows good mixing in function space by effectively capturing multi-modality of complex \glspl{bnn} posterior distribution. This high fidelity in predictive distribution can be attributed to the use of \gls{bma}-loss as meta-objective. 

Also, we conduct 1-D synthetic regression task to visually check whether \gls{l2e} can capture the epistemic uncertainty similar to HMC. \cref{fig:app:synth} shows that \gls{l2e} better captures ``in-between" uncertainty than \gls{de}. Although \gls{hmc} captures epistemic uncertainty more effectively than \gls{l2e}, we observe that \gls{l2e} captures epistemic uncertainty in a similar manner to \gls{hmc}. Please refer to \cref{app:sec:1d} for details.

\subsection{Capturing multi-modality}
In \cref{fig:exp:tsne} and \cref{fig:exp:surface}, we observe the behavior of \gls{de}, \gls{csgmcmc}, and \gls{l2e} in function space. In \cref{fig:exp:surface}, we display the test error surface using a 2-dimensional subspace spanned by the first three collected parameters for each method following \citet{garipov2018loss}. Parameters of \gls{de} clearly located on multiple distinct modes as expected. In contrast, \gls{csgmcmc} seems to sample parameters within a single mode, while samples from \gls{l2e} appear to be in distinct modes. This aligns with the results from \citet{fort2019deep} showing that \gls{csgmcmc} has limited capability of capturing multi-modalities.


For deeper investigation, we plot test error along a linear path between multiple pairs of saved parameters inspired by \citet{goodfellow2014qualitatively}. Existence of loss barrier in \cref{fig:exp:tsne} means the collected parameters belong to different modes. In \cref{fig:exp:tsne}, \gls{l2e} shows a significant increase in predictive error along the linear path between every pair of parameters while \gls{csgmcmc} exhibits a relatively low level of the loss barrier between samples. This suggests that \gls{l2e} is capable of the capturing multi-modality of the posterior distribution. Samples collected from Meta-\gls{sgmcmc} also seem to have loss barriers in CIFAR-10, but considering inferior predictive performance of Meta-\gls{sgmcmc}, the loss barrier among collected parameters from low density region is meaningless.

\begin{table}[t]
    \caption{AUROC for evaluating \gls{ood}. For in-distribution dataset, we use CIFAR-10 and CIFAR-100, while CIFAR-10, CIFAR-100, SVHN, and Tiny-ImageNet serve as \gls{ood} dataset.}
    \label{tab:exp:ood}
    \vskip 0.15in
    \resizebox{\columnwidth}{!}{%
    \begin{tabular}{llrrrrrr}
     \toprule
     In-dist & \gls{ood} & \gls{de} & \gls{csgmcmc} & \gls{l2e} & \gls{hmc} \\
     \midrule
    \multirow{3}{*}{CIFAR-10}
    & CIFAR-100 & 0.844$\spm{0.001}$ & 0.833$\spm{0.002}$ & \textbf{0.857}$\spm{0.001}$ & 0.853\\
    & SVHN & 0.837$\spm{0.005}$ & 0.858$\spm{0.022}$ & \textbf{0.944}$\spm{0.007}$ & 0.892\\
    & {\scriptsize Tiny-ImageNet} & 0.859$\spm{0.001}$ & 0.842$\spm{0.006}$ & \textbf{0.884}$\spm{0.001}$ & 0.849\\
     \midrule
    \multirow{3}{*}{CIFAR-100}
    & CIFAR-10 &  0.726$\spm{0.000}$ & 0.717$\spm{0.001}$ & \textbf{0.743}$\spm{0.002}$ & 0.725\\
    & SVHN & 0.545$\spm{0.013}$ & 0.555$\spm{0.006}$ & 0.786$\spm{0.004}$ & \textbf{0.858}\\
    & {\scriptsize Tiny-ImageNet} & 0.773$\spm{0.001}$ & 0.765$\spm{0.001}$ & \textbf{0.791}$\spm{0.008}$ & 0.770\\
     \bottomrule
    \end{tabular}}
    \vskip -0.2in
\end{table}
\begin{table}[t]
    \centering
     \caption{Accuracy on CIFAR-10-C. We report mean accuracy for 15 types of corruption for each intensity}
     \label{tab:exp:c10c}
     \vskip 0.15in
        \resizebox{0.9\columnwidth}{!}{
        \begin{tabular}{lrrrrrr}
     \toprule
     & \multicolumn{5}{c}{Corruption intensity} \\
     \cmidrule(lr){2-6}
     Method & 1 &  2  & 3 & 4 & 5 & Average
     \\
     \midrule
    \gls{de} & \textbf{0.849} & \textbf{0.824} & \textbf{0.803} & \textbf{0.765} & \textbf{0.714} & 0.791 \\
    \gls{csgmcmc} & 0.835 & 0.807 & 0.784 & 0.744 & 0.694 & 0.773\\
    \gls{l2e} & 0.846 & 0.806 & 0.769 & 0.714 & 0.639 & 0.755\\
    \midrule
    \gls{hmc} & 0.821 & 0.764 & 0.723 & 0.657 & 0.573 & 0.708\\
     \bottomrule
        \end{tabular}}
     \vskip -0.2in
\end{table}

\subsection{\gls{ood} Detection}

We report the \gls{ood} detection performance to estimate the ability to estimate uncertainty. We use \gls{msp} which is equivalent to confidence of logit as \gls{ood} score. We use Area Under the ROC curve(AUROC) \citep{liang2017enhancing} to measure the \gls{ood} detection performance. For Tiny-ImageNet, we resize the image to 32$\times$32 for evaluation. In \cref{tab:exp:ood}, \gls{l2e} shows the best performance regardless of \gls{ood} datasets and in-distribution datasets in general. Only \gls{hmc} outperforms \gls{l2e} on detecting SVHN dataset using neural networks trained on CIFAR-100. One notable result is that the performace gap between  \gls{hmc},\gls{l2e} and other baselines becomes more stark on SVHN. This shows that \gls{hmc} and \gls{l2e} share common features on uncertainty estimation performance and this aligns with other results from \cref{exp:tab:agreement} and \cref{fig:app:synth}.


\begin{table}[t]
    \centering
     \caption{\textbf{Convergence diagonstics}.  \glsunset{ess}\gls{ess} / wall clock time and proportion of parameters with $c\hat{R_{\psi}^{2}} < 1.1$~\citep{sommer2024connecting}. Please refer to \cref{app:sec:converge} for details of metrics.}
     \label{exp:tab:wall}
     \vskip 0.15in
        \resizebox{\columnwidth}{!}{
        \begin{tabular}{llrrrr}
     \toprule
     Metric & Dataset & \gls{csgmcmc}  & Meta-\gls{sgmcmc} & \gls{l2e} 
     \\
     \midrule
    \multirow{4}{4em}{\gls{ess}/s} & \scriptsize fashion MNIST & \textbf{219.85}$\spm{6.64}$ & 33.34$\spm{8.42}$  & 136.31$\spm{0.42}$  \\
    & CIFAR-10 & 56.61$\spm{2.51}$ & 17.31$\spm{5.11}$ & \textbf{82.97}$\spm{0.57}$  \\
    & CIFAR-100 & 41.14$\spm{1.51}$ & 48.29$\spm{7.43}$ & \textbf{63.81}$\spm{0.00}$   \\
    & \scriptsize Tiny-ImageNet & 2.43$\spm{0.01}$ & 0.91$\spm{0.19}$ & \textbf{2.62}$\spm{0.00}$  \\
    \midrule
    \multirow{4}{4em}{\scriptsize $c\hat{R_{\psi}^{2}}<1.1$} & \scriptsize fashion MNIST & 0.765 & 0.524  & \textbf{0.946}  \\
    & CIFAR-10 & 0.804 & 0.160 & \textbf{0.968}  \\
    & CIFAR-100 & 0.732 & 0.750 & \textbf{0.897}   \\
    & \scriptsize Tiny-ImageNet & 0.737 & 0.131 & \textbf{0.806}  \\

     \bottomrule
        \end{tabular}}
     \vskip -0.2in
\end{table}
\subsection{Robustness under covariate shift}

Next, we consider CIFAR-10-C \citep{hendrycks2019benchmarking} for evaluating robustness to covariate shift. In \cref{tab:exp:c10c}, \gls{de} outperforms \gls{l2e} and \gls{csgmcmc} for all intensity levels. \gls{l2e} shows competitive performance under mild corruption, but as the corruption intensifies, the performance of \gls{l2e} significantly drops and shows worst accuracy over all methods. \gls{hmc} also shows this trend, showing worst performance in general. Since \gls{l2e} makes similar predictive distribution with HMC in CIFAR-10, this aligns with the result from ~\citep{izmailov2021dangers,izmailov2021bayesian} that \gls{hmc} and  methods having high fidelity to \gls{hmc} suffer greatly from the covariate shift. Although \gls{l2e} is not robust at covariate shift, it is understandable considering the similarity of \gls{l2e} to \gls{hmc} in function space.



\subsection{Convergence analysis}

We also evaluate sampling efficiency and degree of mixing of \gls{l2e} using \gls{ess} and $c\hat{R_{\psi}^{2}}$~\citep{sommer2024connecting}. Please refer to \cref{app:sec:converge} for details of metrics. In \cref{exp:tab:wall}, \gls{l2e} is the only method that consistently demonstrates decent performance both in terms of \gls{ess} and $c\hat{R_{\psi}^{2}}$ across experiments. On the other hand, other methods show poor mixing, indicating that they hardly explore multi-modal \glspl{bnn} posterior distributions

\glsreset{ess}

\section{Conclusion}
\label{main:sec:conclusion}

In this work, we introduced a novel meta-learning framework called \gls{l2e} to improve \gls{sgmcmc} methods. Unlike conventional \gls{sgmcmc} methods that heavily rely on manually designed components inspired by mathematical or physics principles, we aim to learn critical design components of \gls{sgmcmc} directly from data. Through experiments, we show numerous advantages of \gls{l2e} over existing \gls{sgmcmc} methods, including better mixing, improved prediction performance. Our approach would be a promising direction to solve several challenges that \gls{sgmcmc} methods face in \glspl{bnn}.

\paragraph{Ethics and Reproducibility statement}
Please refer to \cref{app:sec:experiments} for full experimental detail including datasets, models, and evaluation metrics. We have read and adhered to the ethical guideline of International Conference on Machine Learning in the course of conducting this research.

\section*{Acknowledgement}
The authors gratefully acknowledge Giung Nam for helping experiments and constructive discussions. This work was partly supported by  Institute of Information \& communications Technology Planning \& Evaluation (IITP) grant funded by the Korea government (MSIT) (No.RS-2019-II190075, Artificial Intelligence Graduate School Program (KAIST), No.2022-0-00184, Development and Study of AI Technologies to Inexpensively Conform to Evolving Policy on Ethics, and No.2022-0-00713, Meta-learning Applicable to Real-world Problems) and the National Research Foundation of Korea (NRF) grant funded by the Korea government (MSIT) (No. 2022R1A5A708390812). This research was supported with Cloud TPUs from Google's TPU Research Cloud (TRC).

\section*{Impact Statement}
This paper presents work whose goal is to advance the field of Machine Learning. There are many potential societal consequences 
of our work, none which we feel must be specifically highlighted here.

\bibliography{reference}
\bibliographystyle{icml2024}

\newpage
\appendix
\onecolumn

\section{Analysis of exploration property}
\label{app:sec:analysis}

\begin{figure}[ht]
    \centering
    \includegraphics[width=0.9\textwidth]{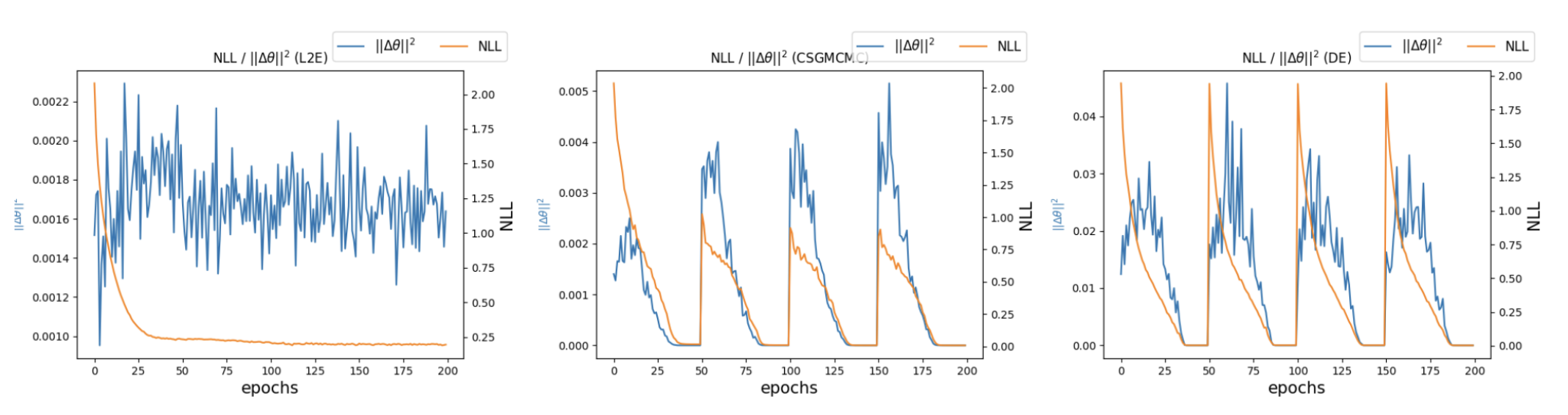}
    \caption{Plots of $||\Delta \theta||^2$ and train \gls{nll} during training of \gls{l2e},\gls{csgmcmc}, \gls{de} on CIFAR-10. Unlike other methods, \gls{l2e} actively updates $\theta$ in the local minima while maintaining training \gls{nll} as nearly constant.}
    \label{fig:app:update}
\end{figure}

\begin{figure}[ht]  
    \centering
    \includegraphics[scale=0.49]{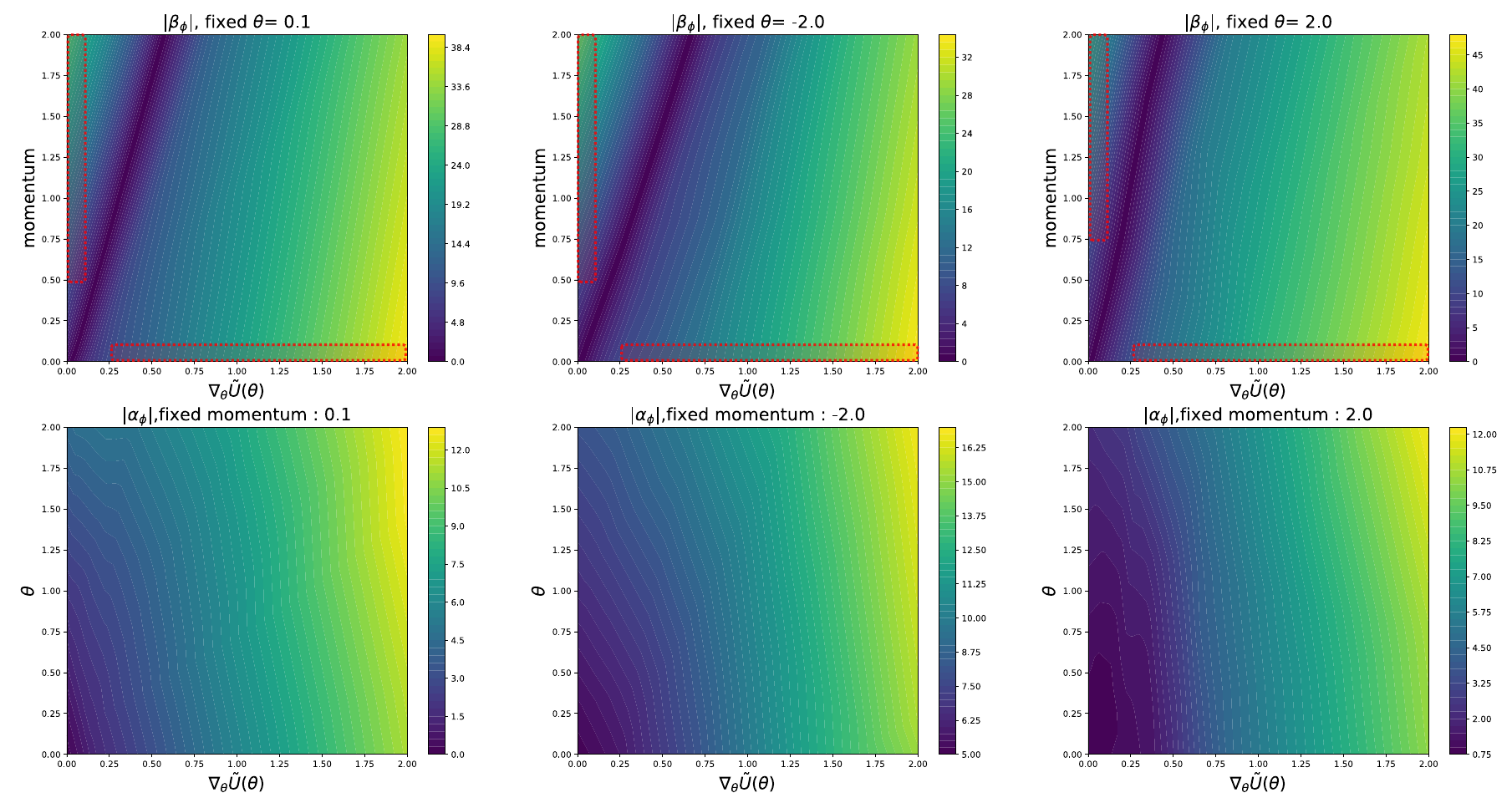}
    \caption{Countour plots of absolute value of outputs of $\beta_\phi$(top) and $\alpha_\phi$(bottom) on the grid. $\beta_{\phi}$ produces large magnitude of output when $\nabla_{\theta}\tilde{U}(\theta)$ is high. When $\nabla_{\theta}\tilde{U}(\theta)$ gets smaller, the overall magnitude decreases as expected, but even when $\nabla_{\theta}\tilde{U}(\theta)$ is nearly zero, $\beta_\phi$ can still allow the sampler to move around posterior distribution when integrated with high momentum value. The regions marked with red dashed boxes can be beneficial for exploration in high density regions. $\alpha_\phi$ is proportional to  $\nabla_{\theta}\tilde{U}(\theta)$ in general, which helps the sampler fastly converge to the high density region.}
    \label{fig:app:contour}
\end{figure}

In this section, we analyze how \gls{l2e} can collect diverse set of parameters with a single MCMC chain. In \cref{fig:app:update}, we analyze the behavior of \gls{l2e} in downstream tasks(CIFAR-10,100) and in \cref{fig:app:contour}, we visualize the scale of outputs of $\alpha_\phi$ and $\beta_\phi$ on the regular grid of input values following \citet{gong2018meta}. Firstly, we plot $l2$ norm of $\Delta \theta = \theta_{t+1} - \theta_t$ at time $t$ and training cross-entropy loss (\gls{nll}) for 200 epochs and comparing it with \gls{de} and \gls{csgmcmc}. Recall the update rule of \gls{l2e}, 
\[
& r_{t+1} = r_t - \epsilon_t [ \nabla_\theta \tilde{U}(\theta_t) + \alpha_\phi(\theta_t, r_t) + C\beta_\phi(\theta_t, r_t)] + \xi_t \\
& \theta_{t+1} = \theta_t + \epsilon_t \beta_\phi(\theta_t, r_{t+1}). \label{eq:update2} \]

where $\xi_t \sim \calN(0, 2C\epsilon_t)$. According to the equation above, $\beta_\phi$ is responsible for updating $\theta$, so tracking $l2$ norm of $\Delta \theta$ is same as tracking $l2$ norm of $\beta_\phi$ since $||\beta_\phi||^2 = \frac{||\Delta\theta||^2}{\epsilon^2}$. 

\begin{figure}[ht]
    \centering
    \includegraphics[width=0.9\textwidth]{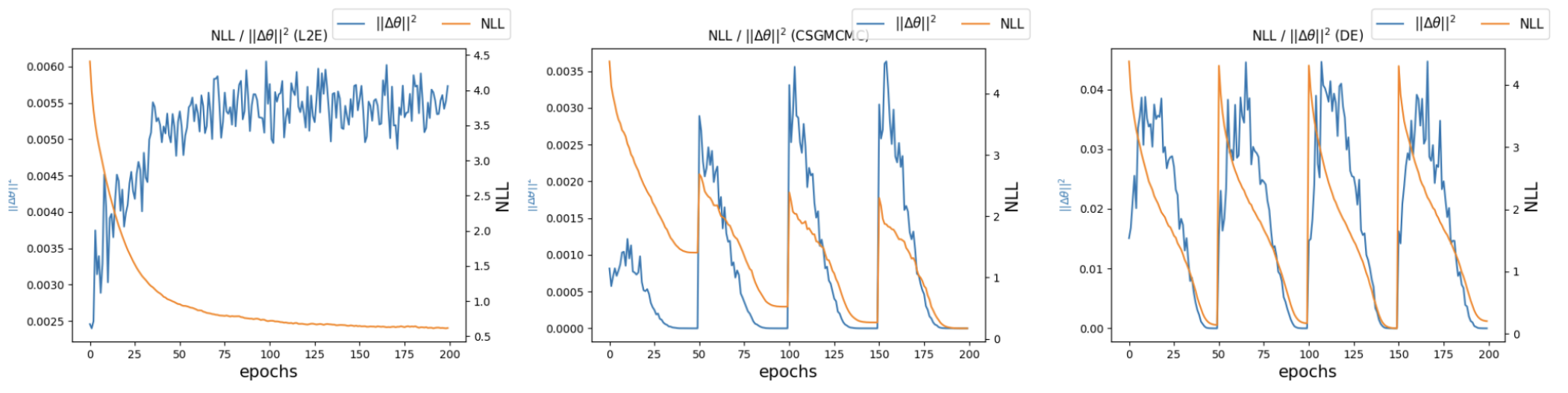}
    \caption{Plots of $||\Delta \theta||^2$ and train \gls{nll} during training of \gls{l2e},\gls{csgmcmc}, \gls{de} on CIFAR-100.}
    \label{fig:app:updatec100}
\end{figure}

In \cref{fig:app:update,fig:app:updatec100}, we find that \gls{l2e} updates $\theta$ with a larger magnitude in local minima than in the early stages of training. This tendency is different from other gradient-based optimizer or \gls{mcmc} methods where the amount of update is relatively small at local minima. Additionally, we notice that L2E actively updates $\theta$ at minima while maintaining loss as nearly constant. This trend is consistently observed in both CIFAR-10 and CIFAR-100, implying that \gls{l2e} learns some common knowledge of posterior information across tasks for efficient exploration in low loss regions. Various experimental results (e.g., see \cref{fig:exp:surface}) support that \gls{l2e} is good at capturing multi-modalities of \glspl{bnn} posterior with a single trajectory. Since \gls{l2e} produces significant amount of updates at local minima without increasing the loss, we can say that our parameterized gradients learned the general knowledge to explore high density regions among different modes.

In \cref{fig:app:contour}, we plot absolute value of outputs of $\alpha_\phi$ and $\beta_\phi$ on the regular input grid. Since $\alpha_\phi$ and $\beta_\phi$ take $\theta,r,\nabla_{\theta}\tilde{U}(\theta)$ and running average of $\nabla_{\theta}\tilde{U}(\theta)$ as inputs, analyzing the function itself is a complex problem. Therefore, to simplify the analysis, we follow the approach of \citet{gong2018meta}, where we fix other inputs except for the statistics we are interested in. For $\beta_{\phi}$, we choose three different fixed values of $\theta$ and plot the results, while for $\alpha_\phi$, we fix the momentum. We assume that there is no running average of $\nabla_{\theta}\tilde{U}(\theta)$ so that running average term is fixed to $\nabla_{\theta}\tilde{U}(\theta)$.

Since the value of $\theta$ itself does not encode the information about the posterior landscape, there is no clear distinction among contour plots in top row with different $\theta$ values. In general, the scale of outputs of  $\alpha_{\phi}$ and $\beta_{\phi}$ is proportional to $\nabla_{\theta}\tilde{U}(\theta)$ which is desirable for fast convergence to high density regions of posterior distribution. For $\beta_{\phi}$,  when $\nabla_{\theta}\tilde{U}(\theta)$ gets smaller, the overall magnitude of output decreases as expected, but even when $\nabla_{\theta}\tilde{U}(\theta)$ is nearly zero, $\beta_\phi$ can still make large magnitude of update of $\theta$ when integrated with high momentum value. Also, when momentum is nearly zero, it still allows sampler to explore posterior distribution when $\nabla_{\theta}\tilde{U}(\theta)$ is far from zero. This complementary relationship between two statistics is strength of \gls{l2e} that the movement of sampler is not solely depend on a single statistic so that it can exploit more complex information about loss geometry unlike standard SGHMC. $\alpha_{\phi}$ produces an output proportional to the scale of the gradient regardless of the momentum values. This implies that $\alpha_{\phi}$ helps the acceleration of sampler in low-density regions as it is added to the energy gradient in \cref{eq:update2}.

One limitation of our analysis is that we focus on analyzing the magnitude of the function output rather than its direction. Although the magnitude of $\alpha_\phi$ and $\beta_\phi$ are closely connected to the exploration property, delving deeper into how L2E traverses complex and multi-modal \glspl{bnn} posterior landscape would be an interesting direction of future research.

\section{Bayesian Model Average(BMA) vs Cross-Entropy(CE) meta-loss}
\label{app:sec:metaloss}

\begin{figure}[ht]
    \centering
    \includegraphics[width=0.6\textwidth]{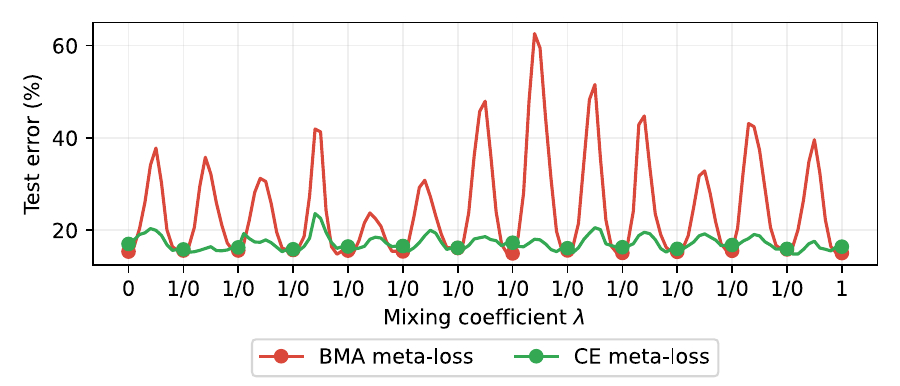}
    \caption{Ablation study of BMA vs CE meta-loss on CIFAR-10.}
    \label{fig:app:bmace}
\end{figure}

In this section, we will explain why \gls{bma} meta-loss enhances the exploration of the sampler, by comparing it with CE meta-loss. Average CE loss of individual models along the optimizer's trajectory on meta-training tasks has been used as a meta-objective to train learned optimizers~\citep{andrychowicz2016learning, metz2019understanding, metz2022velo} on classification tasks. Two different objectives have following forms:

\[
L^{\text{CE}}(\phi) = - \frac{1}{K}\sum_{k=1}^K \log p(y_*\given x_*, \theta_k(\phi)) \\
L^{\text{\gls{bma}}}(\phi) = - \log \frac{1}{K}\sum_{k=1}^K p(y_*\given x_*, \theta_k(\phi))
\]
where $(x_*, y_*)$ is a validation data point.  While \gls{bma} meta-loss is similar to CE meta-loss, \gls{bma} meta-loss differs significantly from CE meta-loss. \gls{bma} meta-loss is Monte Carlo approximation of the posterior predictive distribution for validation data points. Actually, we gather models along the sampler's trajectory and minimize CE-loss of the average probability of collected models. \gls{bma} meta-loss not only encourages the sampler to minimize the average CE-loss of individual models but also promotes increased functional diversity among collected models. According to \citet[Equation 7]{wood2023unified}, this loss of the ensemble model is decomposed as “ensemble loss = average individual loss - ambiguity”. Ambiguity refers to the difference among the ensemble models and individual models, and the larger it is, the more \gls{bma} meta-loss decreases. In \cref{fig:app:bmace}, we observe that \gls{l2e} trained with \gls{bma} meta-loss exhibits a much greater loss barrier between collected parameters than \gls{l2e} trained with CE meta-loss, indicating the higher exploration and larger functional diversity among samples. Also, our extensive downstream experiments and visualization demonstrates that L2E with \gls{bma} meta-loss actually increases functional diversity among collected models.

\section{\gls{l2e} on Text Dataset}
\label{app:sec:imdb}
In order to check whether \gls{l2e} can adapt well to the other modalities (e.g. text dataset), we additionally conduct text classification on IMDB dataset with CNN-LSTM architecture following \citet{wenzel2020good} and \citet{izmailov2021bayesian}. 

For \gls{hmc}, we use checkpoints of 1200 samples from \citet{izmailov2021bayesian}. For other MCMC methods, we collect 50 samples with 50 epochs of thinning interval with 100 burnin epochs. \cref{app:tab:kineticl2e_right} demonstrates that \gls{l2e} shows competitive performance on the text dataset, meaning that \gls{l2e} can still work well on unseen modalities. This transfer of knowledge from image datasets to text dataset supports that the generalization capacity of \gls{l2e} is strong compared to \citet{gong2018meta}, which is only proven to generalize well to similar datasets with meta-training dataset.

\section{Results on alternative parameterization}
\label{app:sec:kineticl2e}
\begin{figure*}[t]
\centering
\begin{subfigure}{0.28\textwidth}
    \includegraphics[width=\textwidth]{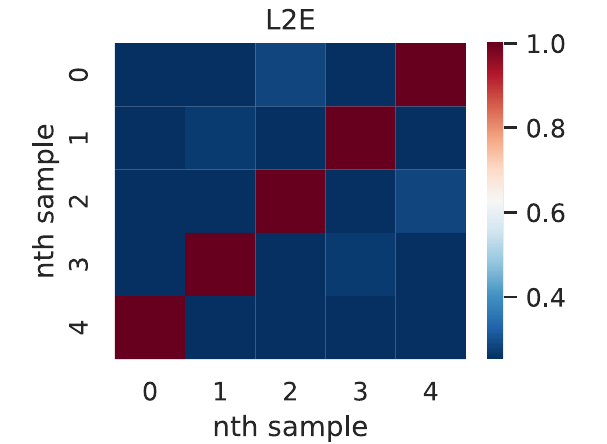}
    \caption{Cosine similarity between weights.}
    \label{fig:app:param_cosine}
\end{subfigure}
\hfill
\begin{subfigure}{0.28\textwidth}
    \includegraphics[width=\textwidth]{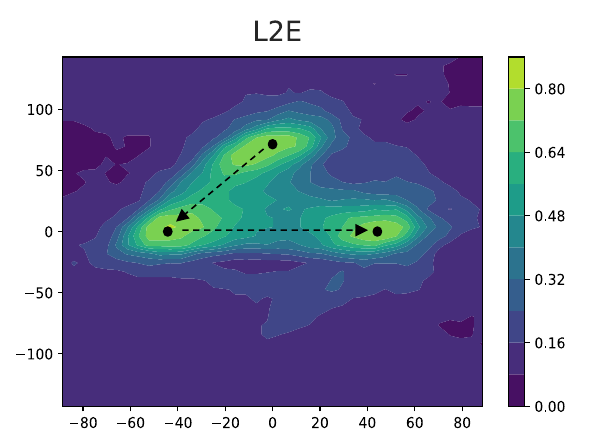}
    \caption{Test error ($\%$) around the samples.}
    \label{fig:app:param_surface}
\end{subfigure}
\hfill
\begin{subfigure}{0.42\textwidth}
    \includegraphics[width=\textwidth]{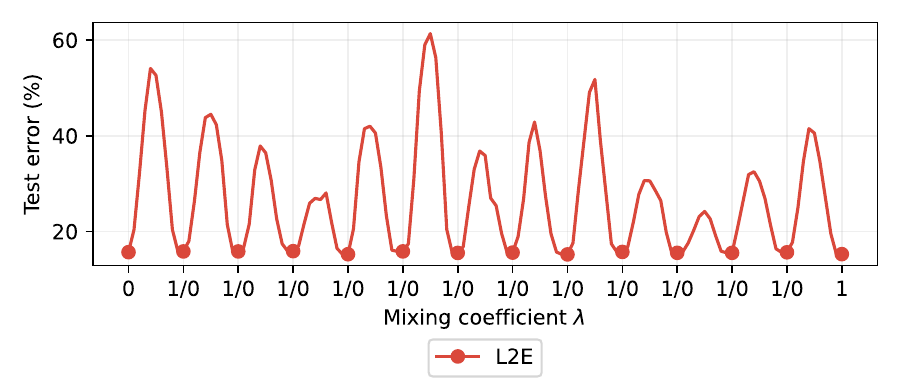}
    \caption{Test error ($\%$) along linear path}
    \label{fig:app:param_test}
\end{subfigure}
\caption{Figures show multi-modality of \gls{l2e} with parameterization of momentum distribution. (a) shows cosine similarity between weights. (b) is loss surface as a function of model parameters in a 2-dimensional subspace spanned by solutions of each method. Colors represent the level of test error. (c) shows test error ($\%$) along linear path between a pair of parameters.}
\label{fig:exp:multi_modality_kinetic}
\end{figure*}

Although parameterizing $\nabla g(\theta,r)$ allows learned sampler to be expressive and efficient, we should make assumptions on the underlying $g(\theta,r)$. To avoid introducing additional assumptions, we propose another version of \gls{l2e} that directly parameterizes kinetic energy function $g(\theta,r)$ rather than its gradients and evaluate whether it can achieve comparable performance comparable to \gls{l2e}. Specifically, we fix $p(r|\theta)$ as normal distribution and parameterize its mean with $f_\phi(\theta)$ that takes $\theta,\nabla_\theta \tilde{U}(\theta)$ as inputs. This approach eliminates the need for assumptions regarding the existence and integrability of unnormalized probability density function. Since we set $p(r|\theta) \sim \mathcal{N}(f_\phi(\theta),I)$, kinetic energy function of $p(r|\theta)$ is $g(\theta,r) = (r-f_\phi(\theta))^T(r-f_\phi(\theta))$. Therefore, we can get $\nabla_{\theta}g(\theta,r) = f_\phi(\theta) \nabla_{\theta}f_\phi(\theta)$ and  $\nabla_{r}g(\theta,r) = r-f_\phi(\theta)$. Then, update rule for this parameterization is as follows
\[
& r_{t+1} = r_t - \epsilon_t (\nabla_\theta \tilde{U}(\theta_t) + f_\phi(\theta) \nabla_{\theta}f_\phi(\theta) + \alpha (r_t-f_\phi(\theta)) +\xi_t  \\
& \theta_{t+1} = \theta_t + \epsilon_t (r_t - f_\phi(\theta))  
\label{eq:kineticenergy} \]
where $\xi_t \sim \calN(0, 2\epsilon_t\alpha)$. We will refer to this version of sampler as Kinetic-\gls{l2e}. In \cref{app:tab:kineticl2e_left}, Kinetic-\gls{l2e} shows comparable performance to \gls{l2e} across all image classification experiments. Also, \cref{fig:exp:multi_modality_kinetic} demonstrates the exploration capacity of Kinetc-\gls{l2e}. These results indicate that both versions can achieve similar performance in terms of predictive accuracy and exploration.

Both versions of \gls{l2e} have their own strength and weakness. \gls{l2e} requires assumptions on the underlying kinetic energy function to ensure the convergence guarantee by \citet{ma2015complete}, but it is computationally efficient and can model complex functions. Kinetic-\gls{l2e} guarantee the convergence without additional assumptions on $g(\theta,r)$, but requires additional gradient computations and conditional distribution $p(r|\theta)$ is restricted to specific distribution which can possibly harm the flexibility of the sampler. We recommend to consider these trade-offs for further applications. 

 

 


 


 

\begin{minipage}[t]{0.55\textwidth}
    \centering
    \captionof{table}{Results of kinetic energy parameterization of \gls{l2e}.}
    \vskip 0.15in
    \resizebox{\linewidth}{!}{
    \begin{tabular}{lrrrrr}
        \toprule
        Dataset & \gls{l2e} & ACC $\uparrow$ & NLL $\downarrow$ & ECE $\downarrow$  & KLD $\uparrow$ \\
        \midrule
        \multirow{2}{*}{fashion-MNIST}
        & Kinetic \gls{l2e} & 0.9175 & 0.2425 & 0.0104 & 0.1334  \\
        & \gls{l2e} & 0.9166 & 0.2408 & 0.0078 & 0.0507  \\
        \midrule
        \multirow{2}{*}{CIFAR-10}
        & Kinetic \gls{l2e} & 0.9131 & 0.2867 & 0.0527 & 0.4152  \\
        & \gls{l2e} & 0.9123 & 0.2909 & 0.0563 & 0.4344 \\
        \midrule
        \multirow{2}{*}{CIFAR-100}
        & Kinetic \gls{l2e} & 0.7013 & 1.114 & 0.1212 & 1.735 \\
        & \gls{l2e} & 0.6999 & 1.131 & 0.1403 & 1.749 \\
        \midrule
        \multirow{2}{*}{Tiny-ImageNet}
        & Kinetic \gls{l2e}  & 0.5561  & 1.966 & 0.1731 & 1.907  \\
        & \gls{l2e} & 0.5583 & 1.846 & 0.1423 & 0.9139 \\
        \bottomrule
    \end{tabular}
    }
    \label{app:tab:kineticl2e_left}
    \vskip -0.1in
\end{minipage}%
\begin{minipage}[t]{0.40\textwidth}
    \centering
    \captionof{table}{Results on IMDB classification task, following the experimental setup of \citet{izmailov2021bayesian}}
    \vskip 0.15in
    \resizebox{\linewidth}{!}{
    \begin{tabular}{lrr}
        \toprule
        Method & \multicolumn{1}{c}{ACC $\uparrow$} & \multicolumn{1}{c}{NLL $\downarrow$} \\
        \midrule
        \gls{de} & 0.867$\spm{0.000}$ & 0.386$\spm{0.000}$ \\
        \midrule
        \gls{csgmcmc} & 0.848$\spm{0.007}$ & 0.401$\spm{0.032}$ \\
        \midrule
        \gls{hmc} & 0.868$\spm{0.000}$ & 0.308$\spm{0.000}$\\
        \midrule
        Meta-\gls{sgmcmc} & 0.820$\spm{0.003}$ & 0.401$\spm{0.001}$\\
        \midrule
        \gls{l2e} & \textbf{0.873}$\spm{0.001}$ & \textbf{0.301}$\spm{0.001}$\\

        \bottomrule
    \end{tabular}
    }
    \label{app:tab:kineticl2e_right}
    \vskip -0.1in
\end{minipage}%

\section{Comparision with \citet{gong2018meta}}
\label{app:sec:metasgmcmc}





In~\cref{tab:metavl2e}, we demonstrate the difference between Meta-\gls{sgmcmc}~\citep{gong2018meta} and \gls{l2e}. Meta-\gls{sgmcmc} aims to build a sampler through meta-learning that rapidly converges to the target distribution and performs accurate simulation. This goal aligns with the objectives of all \gls{sgmcmc} methods. However, our approach is specifically designed with the more concrete purpose of effectively simulating multi-modal DNN posterior distribution and also generalizing to unseen problems. We compare \gls{l2e} to Meta-\gls{sgmcmc} in various aspects in the following subsections.

\def\textBF#1{\sbox\CBox{#1}\resizebox{\wd\CBox}{\ht\CBox}{\textbf{#1}}}

\glsunset{tbptt}

\begin{table}[t]
    \centering
    \caption{Comparison between \citet{gong2018meta} and L2E}
    \vskip 0.15in
  \label{tab:metavl2e}
    \resizebox{\linewidth}{!}{
    \begin{tabular}{ccc}
 \toprule
  & \citet{gong2018meta} & \gls{l2e} \\

 \midrule
\midrule

Purpose of meta-learning & Fast convergence, low bias & Efficient exploration of multi-modal \glspl{bnn} posterior \\
\midrule
Learning target & Diffusion and curl matrix & Gradient of kinetic energy \\
\midrule
Meta-training task & Single-task & Multi-task  \\
\midrule
Meta-objective & $\textrm{KL}(q_t | \pi)$ & $- \log \frac{1}{K} \sum_{k=1}^K p(y_*\given x_*, \theta_k(\phi))$  \\
\midrule
Meta-gradient estimation & \gls{tbptt}~\citep{werbos1990backpropagation} & \gls{es}~\citep{salimans2017evolution} \\
\midrule
Generalizes to & unseen classes in the same dataset & completely different datasets \\
\midrule
Scales to & small scale architecture & large scale architecture\\

 \bottomrule
    \end{tabular}}
\end{table}

\begin{table}[ht]
    \centering
    \caption{Experiments on CIFAR-10 following experiments in \citet{gong2018meta}. Without having seen CIFAR-10 dataset during meta training, \gls{l2e} outperforms reported performance of Meta-\gls{sgmcmc} while using significantly less computational cost than Meta-\gls{sgmcmc}}. 
    \resizebox{1.0\linewidth}{!}{
    \begin{tabular}{lllllll}
 \toprule
 Methods & NT ACC & NT+AF ACC & NT+Data ACC & NT NLL/100 & NT+AF NLL/100 & NT+Data NLL/100  \\
 \midrule
 
Meta-\gls{sgmcmc} & 78.12$\spm{0.035}$ & 74.41$\spm{0.11}$ & 89.97$\spm{0.04}$ & 68.88$\spm{0.15}$ & 79.55$\spm{0.057}$ & 15.66$\spm{0.28}$ \\
\gls{l2e} & \textbf{79.21}$\spm{0.203}$ & \textbf{75.91}$\spm{0.200}$ & \textbf{92.49}$\spm{0.234}$ & \textbf{63.23}$\spm{0.46}$ & \textbf{72.11}$\spm{0.22}$ & \textbf{11.82}$\spm{0.81}$\\
 \bottomrule
     \label{metavl2ecifar}

    \end{tabular}}
\end{table}

\begin{table}[ht]
    \centering
    \caption{Experiments on MNIST following experiments in \citet{gong2018meta}}.
    \resizebox{1.0\linewidth}{!}{
    \begin{tabular}{lllllll}
 \toprule
 Methods & NT ACC & NT+AF ACC & NT+Data ACC & NT NLL/100 & NT+AF NLL/100 & NT+Data NLL/100  \\
 \midrule
 
Meta-\gls{sgmcmc} & 98.36$\spm{0.02}$ & 97.72$\spm{0.02}$ & 98.62$\spm{0.02}$ & 640$\spm{6.25}$ & 875$\spm{3.19}$ & 230$\spm{3.23}$ \\
\gls{l2e} & \textbf{98.39}$\spm{0.05}$ & \textbf{98.07}$\spm{0.08}$ & - & \textbf{558}$\spm{3.19}$ & \textbf{679}$\spm{6.65}$ & - \\

 \bottomrule
        \label{metavl2emnist}
    \end{tabular}}
\end{table}

\subsection{Difference in parameterization of meta models}
The update rule for $z = (\theta, r) \in \bbR^{2d}$ presented in~\citet{gong2018meta} is as following. 
\[
& r_{t+1} = (1-\epsilon_t D_f(z_t))r_t - \epsilon_t Q_f(z_t)\nabla_\theta \tilde{U}(\theta_t) + \epsilon_t \Gamma_r(z_t) +\xi_t  \\
& \theta_{t+1} = \theta_t + \epsilon_t Q_f(z_t) r_t + \epsilon_t \Gamma_\theta(z_t) \\
& \xi_t \sim \calN(0, 2\epsilon_tD_f(z_t)), \quad \Gamma_i(z) := \sum_{j=1}^{2d} \frac{\partial}{\partial z_j}(D_{ij}(z) + Q_{ij}(z))
\label{eq:metasgmcmc}
\]



In our main text, we point out that parameterizing $D_f$ and $Q_f$ makes additional computational burden. Additionally, since $D_f$ and $Q_f$ mainly function as multipliers for gradient and momentum, learning them may not be as effective as it should be for effective exploration in low energy regions. In low energy regions where the norm of gradient and momentum are extremely small, it is difficult to make reasonable amount of update of $\theta$ for exploration by multiplying $Q_f$ to momentum and gradient. Also, $Q_f$ should be clipped by some threshold for practical issue. By contrast, In the update rule of \gls{l2e} \eqref{eq:update}, $\alpha_\phi , \beta_\phi$ are added to the gradient, which is more suitable for controlling the magnitude and direction of update in low energy regions. Also, ablation study in~\cref{app:tab:ablation} shows the inferior performance of learning $D_f$ and $Q_f$ in terms of classification accuracy and predictive diversity in our setting.
Therefore, we choose to learning kinetic energy gradient is better than learning $D_f$ and $Q_f$, as it avoids additional computation and allows the sampler to mix better especially around the low energy region. 

 


 


\subsection{Difference in meta-training procedure}
\glsreset{tbptt}
There are significant differences between two methods in meta-training pipeline. Firstly, the most notable distinction is that Meta-\gls{sgmcmc} learns with only one task, requiring the training of a new learner for each problem. This poses an issue as a single learner may not generally apply well to various problems. In contrast, our approach involves sampling from a task set composed of diverse datasets and architectures for training. Another crucial difference is the choice of estimator for the meta-objective gradients. \citet{gong2018meta} employs \gls{tbptt} which truncates computational graphs of a long inner-loop to estimate the meta gradient. While this approach can save computational cost by avoiding backpropagation through long computational graph, it results in a biased estimator of meta-gradient. To address these issues, we utilize \gls{es} to compute unbiased estimator of meta-gradient efficiently.

\subsection{Experimental results}

In order to compare \gls{l2e} with Meta-\gls{sgmcmc}, we exactly replicate the experimental setup of MNIST and CIFAR-10 experiments in \citet{gong2018meta} and evaluate performance of \gls{l2e}. These experiments evaluate how well each method generalizes to unseen Neural Network architecture (NT), activation function (AF) and dataset (Data) which were unseen during the meta-learning process. For the experimental details, please refer to the experiments section of \citet{gong2018meta}. We use same scale of metrics in \citet{gong2018meta} for conveniently comparing two methods. We report \gls{acc} and \gls{nll} on test dataset. Since \gls{l2e} used MNIST dataset during meta-train, we do not evaluate Dataset generalization experiments on MNIST.  Meta-\gls{sgmcmc} used 20 parallel chains with 100 epochs for MNIST and 200 epochs for CIFAR-10. We use single chain with 100 burn-in epochs for both experiments, and use 10 thinning epochs for collecting 20 samples. 

In \cref{metavl2ecifar} and \cref{metavl2emnist}, we confirm that \gls{l2e} outperforms Meta-\gls{sgmcmc} for all generalization types in terms of \gls{acc} and \gls{nll} despite using significantly less computational cost. Notably, in CIFAR-10 experiment, despite \gls{l2e} was not trained on the CIFAR-10 during meta-learning, \gls{l2e} significantly outperforms Meta-SGMCMC with a wide margin indicating that our approach better generalizes to unseen datasets compared to Meta-\gls{sgmcmc}.

\section{Discussion on meta-training objectives}

Previous studies have proposed various meta-objectives to achieve goals similar to ours.  \citet{gong2018meta} minimizes the $\textrm{KL}(q_t | \pi)$, where $\pi$ is target distribution and $q_t$ is the marginal distribution of $\theta$ at time $t$ for good mixing. On the other hand, \citet{levy2017generalizing}  employs meta-objective maximizing the jump distances between samples in weight space and simultaneously minimizing the energy in order to make sampler rapidly explore between modes. However, explicitly maximizing the jump distances in weight space can be easily cheated, as the distances between weights does not necessarily lead to the difference in the functions, resulting in trivial sampler with which achieving the balance between convergence and exploration is hard. Also, minimizing the divergence with target distribution seems sensible, but due to intractable $q_t$, computing the gradient of $q_t$ should resort to gradient estimator. \citet{gong2018meta} used stein-gradient estimator, which requires multiple independent chains so it harms scalability. Also, this objective does not lead the learned sampler to explore multi-modal distribution. \citet[Figure 3]{gong2018meta} shows that the learned sampler quickly converges to low energy region, but learned friction term $D_f$ restricts the amount of update in low energy region, limiting the exploration behaviour.  Among choices, we find out that BMA meta-loss is a simple yet effective meta-objective that naturally encodes exploration-exploitation balance without numerical instability and exhaustive hyperparameter tuning.

\begin{figure}[t]
    \centering
    \includegraphics[width=\textwidth]{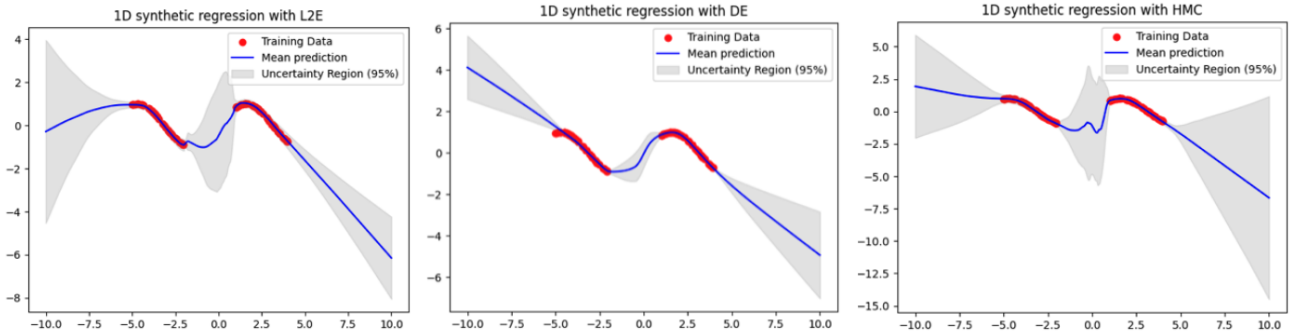}
    \caption{1-D synthetic regression of \gls{l2e}, \gls{de}, \gls{hmc}. See \cref{app:sec:1d} for details and discussion.}
    \label{fig:app:synth}
\end{figure}

\section{1-D synthetic regression}
\label{app:sec:1d}

We conduct 1-D synthetic regression task to visually check whether \gls{l2e} can capture the epistemic uncertainty. For the training data, we generate 1000 data points from underlying true function $y=\sin(x)$, within the interval $[-5,1]$ and $[1,4]$. We use \gls{de} and \gls{hmc} as baselines. We collect 50 parameters for each methods and plot the mean prediction and $95\%$ confidence interval of the prediction. For \gls{l2e}, we do not fine-tune the learned sampler used in the main experiments. We use thinning interval of 50 training steps and 1000 burn-in steps for \gls{l2e}, 300 training steps for each single solution of \gls{de} and 1000 burn-in steps and 100 leap-frog steps for \gls{hmc}. We use 2 layers MLP with 100 hidden units and ReLU activation to estimate the function. 

Effective method for capturing epistemic uncertainty should make confident predictions for the training data and should be uncertain on \gls{ood} data points. \cref{fig:app:synth} shows that \gls{l2e} better captures 'in-between' uncertainty than \gls{de}. \gls{l2e} generally produces diverse predictions for out-of-distribution data points, especially for the input space between two clusters of training data while \gls{de} shows relatively confident prediction in that region. \gls{hmc} is known for the golden standard for posterior inference in Bayesian method. Our experimental result aligns with this common knowledge since \gls{hmc} is the best in terms of capturing epistemic uncertainty among three methods especially in areas out of the range of training points. While our approach falls short of \gls{hmc}, it demonstrates better uncertainty estimation than that of \gls{de} and shows similar predictive uncertainty with \gls{hmc} between two training points cluster. These experimental results align with the \gls{ood} detection experiments and convergence diagnostics presented in our main text, indicating \gls{l2e} performs effective posterior inference. It is important to note that even without meta-training on the regression datasets, \gls{l2e} can adapt well to the regression problem.

\section{Image classification with Data Augmentation}
\label{app:sec:aug}

Since applying data augmentation violates Independently and Identically Distributed(IID) assumption of the dataset which is commonly assumed by Bayesian methods, this can lead to model misspecification~\citep{wenzel2020good,kapoor2022uncertainty} and under model misspecifcation, bayesian posterior may not be the optimal for the \gls{bma} performance~\citep{masegosa2020learning}. Therefore, prior work such as \citet{izmailov2021bayesian} argued the incompatibility between Bayesian methods and data augmentation. However, data augmentation is an indispensable technique in modern machine learning so it is also interesting to see how \gls{l2e} performs with data augmentation. We run the image classification experiments on CIFAR-10 and CIFAR-100 with random crop and horizontal flip for the data augmentation. Since using data augmentation introduces Cold Posterior Effect~\citep{wenzel2020good} for Bayesian methods such as \gls{csgmcmc} and \gls{l2e}, we additionally tune the temperature hyperparameter for these methods. We tune the temperature for each methods, using $T=0.0001$ in both experiments. Please refer to ~\citet{wenzel2020good} for detailed analysis of Cold Posterior Effect. We collect 10 samples for all methods for these experiments.

In~\cref{tab:exp:clsaug}, when applying data augmentation, the performance gap between \gls{csgmcmc} and \gls{l2e} becomes far less stark than without data augmentation. Moreover, \gls{de} outperforms other baselines with a large margin in CIFAR-10 and shows similar results with \gls{l2e} in CIFAR-100 since \gls{l2e} significantly outperforms other method without data augmentation. With data augmentation, it is not very surprising that \gls{de} outperforms other Bayesian methods like \gls{l2e} and \gls{csgmcmc} in terms of predictive accuracy and calibration since they suffer from model misspecification and temperature tuning can partially handle this problem~\citep{kapoor2022uncertainty}. Observing the high agreement between \gls{l2e} and \gls{hmc} in the function space suggests that \gls{l2e} effectively approximates the predictive distribution of the target posterior distribution. However, when techniques that violate assumptions for model likelihood are applied, correctly simulating target distribution does not necessarily mean the superior performance than other methods. As a result, there may be a reduction or even a reversal in the performance gap compared to other methods. Nevertheless, \gls{l2e} demonstrates comparable performance to DE with less compute on datasets like CIFAR-100. Also, \gls{l2e} shows slightly better performance than Bayesian method,\gls{csgmcmc} in CIFAR-10 and outperforms with a wide margin in CIFAR-100 experiment.  When it comes to predictive diversity, \gls{l2e} significantly outperforms baselines on both experiments. Although applying data augmentation introduces significant variations to the posterior landscape, we confirm that \gls{l2e} still maintains the exploration property. To sum up, we argue that \gls{l2e} is still practical method even with data augmentation since it shows competitive predictive performance and efficiently explores the posterior landscape.

\def\textBF#1{\sbox\CBox{#1}\resizebox{\wd\CBox}{\ht\CBox}{\textbf{#1}}}


\begin{table}[t]
\centering
   \centering
   \caption{{Results of image classification with data augmentation.}} 
   \resizebox{0.65\textwidth}{!}{
     \setlength{\tabcolsep}{3pt}
\begin{tabular}[t]{L{1.7cm}rrrrr}
 \toprule
 Dataset & Method & ACC $\uparrow$ & NLL $\downarrow$ & ECE $\downarrow$  & KLD $\uparrow$ \\

 \midrule
 
\multirow{3}{*}{CIFAR-10}
& \gls{de} & \textbf{0.931}$\spm{0.002}$ & \textbf{0.209}$\spm{0.003}$ &
\textbf{0.017}$\spm{0.001}$ & 0.293$\spm{0.002}$  \\
& \gls{csgmcmc} & 0.923$\spm{0.003}$ & 0.234$\spm{0.007}$ &
0.031$\spm{0.005}$ & 0.142$\spm{0.019}$  \\
& \gls{l2e} & 0.926$\spm{0.001}$ & 0.235$\spm{0.002}$ &
0.037$\spm{0.001}$ & \textbf{0.391}$\spm{0.005}$  \\

 \midrule
 
\multirow{3}{*}{CIFAR-100}
& \gls{de} & \textbf{0.708}$\spm{0.001}$ & \textbf{1.048}$\spm{0.001}$ &
0.092$\spm{0.001}$ & 0.765$\spm{0.003}$  \\
& \gls{csgmcmc} & 0.683$\spm{0.001}$ & 1.114$\spm{0.005}$ &
\textbf{0.013}$\spm{0.001}$ & 0.319$\spm{0.008}$  \\
& \gls{l2e} & 0.705$\spm{0.003}$ & 1.066$\spm{0.001}$ &
0.095$\spm{0.003}$ & \textbf{0.996}$\spm{0.017}$  \\

 \bottomrule
    \end{tabular}} 
\label{tab:exp:clsaug}
\end{table}

\begin{figure}[ht]
    \centering
    \includegraphics[width=\textwidth]{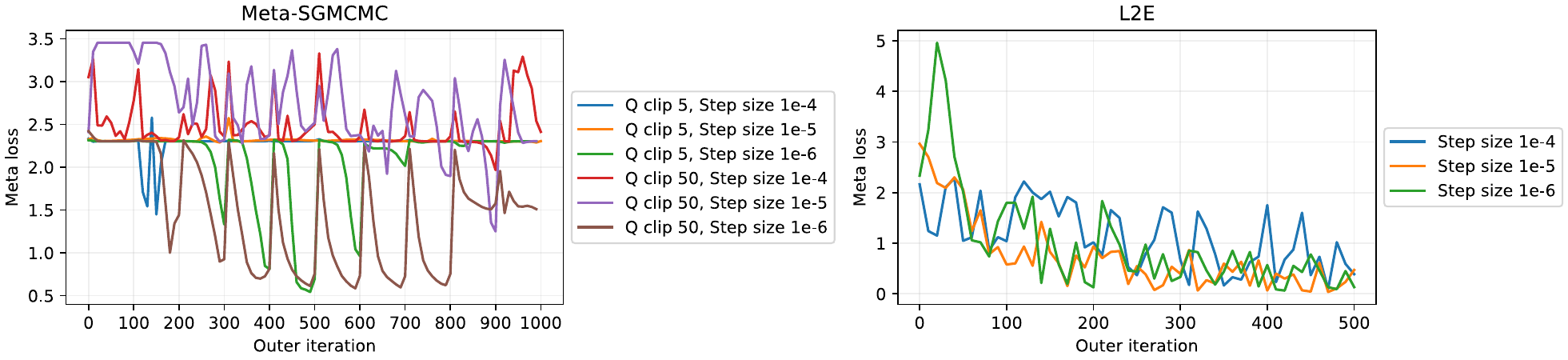}
    \caption{Meta loss at each outer iteration during the meta training process. For Meta-\gls{sgmcmc}, it is evident that the meta loss is highly unstable depending on hyperparameters such as Q clip and step size. Conversely, with \gls{l2e}, it can be observed that the meta loss converges stably even with varying step size values.}
    \label{fig:app:meta_loss}
\end{figure}

\section{Details for meta-training}
For meta-training, we construct our experiment code based on JAX learned-optimization package~\citep{metz2022practical}. Also, we build our own meta-loss and \gls{l2e} with specific input features and design choice. Therefore, we construct our own meta-learning task distribution that \gls{l2e} can efficiently learn knowledge for better generalization for large-scale image classification task.

\label{app:sec:meta}
\subsection{Input features}
We use the following input features for \gls{l2e}:
\begin{itemize}
    \item raw gradient values
    \item raw parameter values
    \item raw momentum values
    \item running average of gradient values
\end{itemize}
Running average feature is expanded for multiple time scale in that we use multiple momentum-decay values for averaging. We use 0.1, 0.5, 0.9, 0.99, 0.999 and 0.9999 for momentum decay so that running average feature is expanded into 6-dimensions. Therefore, we have total 9-dimensional input features for each dimension of parameter and momentum. Input features are normalized so that $l2$ norm with respect to input features of different dimensions become 1. $\alpha_{\phi}$ and $\beta_\phi$ share weights of neural network except for the last layer of \gls{mlp}, so that we can get two quantities with single forward pass. This weight sharing method is employed in \citet{levy2017generalizing} or other recent literature in learned optimization like in \citet{metz2022velo} and \citet{metz2019understanding}.

\begin{algorithm}[t]

    \begin{algorithmic}[1]
    \caption{InnerLoop}\label{alg:inner}
    \STATE {\bfseries Input:} Meta parameter $\phi$, inner iterations $N_\text{inner}$, initial parameter $\theta_0$, step size $\varepsilon$, burn-in steps $B$, thinning interval $T$.
    \STATE {\bfseries Output:} Loss $L(\phi)$
    \STATE Initialize $\Theta = \varnothing$ and $r_0 \sim \calN(0,I_d)$.
    \FOR{$i=1,\dots, N_{\normalfont\text{inner}}$}
    \STATE $r_{t+1} = r_{t} - \epsilon_t (\nabla \tilde{U}(\theta_t) + \alpha_{\phi} + c\beta_{\phi}) + \xi_t$ where $\xi_t \sim \calN(0, 2c)$.
    \STATE $\theta_{t+1} = \theta_t + \epsilon_t \beta_\phi$
    \IF{$i > B \; \& \mod(i,T) =0 $} 
    \STATE $\Theta \gets \Theta \cup \{ \theta_i \}$
    \ENDIF
    \ENDFOR
    \STATE $L(\phi) \gets - \log \frac{1}{|\Theta|}\sum_{\theta \in \Theta} p(y^*|x^*,\theta)$
    \end{algorithmic}   
\end{algorithm}

\subsection{Task distribution}

\paragraph{Dataset} We use MNIST, Fashion-MNIST, EMNIST and MedMNIST for meta training. We do not use resized version of dataset for meta training. For MedMNIST, we use the BloodMNIST in the official website.
\paragraph{Neural network architecture}At each outer iteration, we randomly sample one configuration of neural network architecture which is constructed by the possible choice below. We use the following options for neural network architecture variation:
\begin{itemize}
    \item Size of the convolution output channel : $\{4,8,16\}$
    \item Number of convolution layers: $\{1,2,3,4,5\}$
    \item Presence of the residual connection: boolean
\end{itemize}


\begin{figure}[ht]
    \centering
    \includegraphics[width=0.7\linewidth]{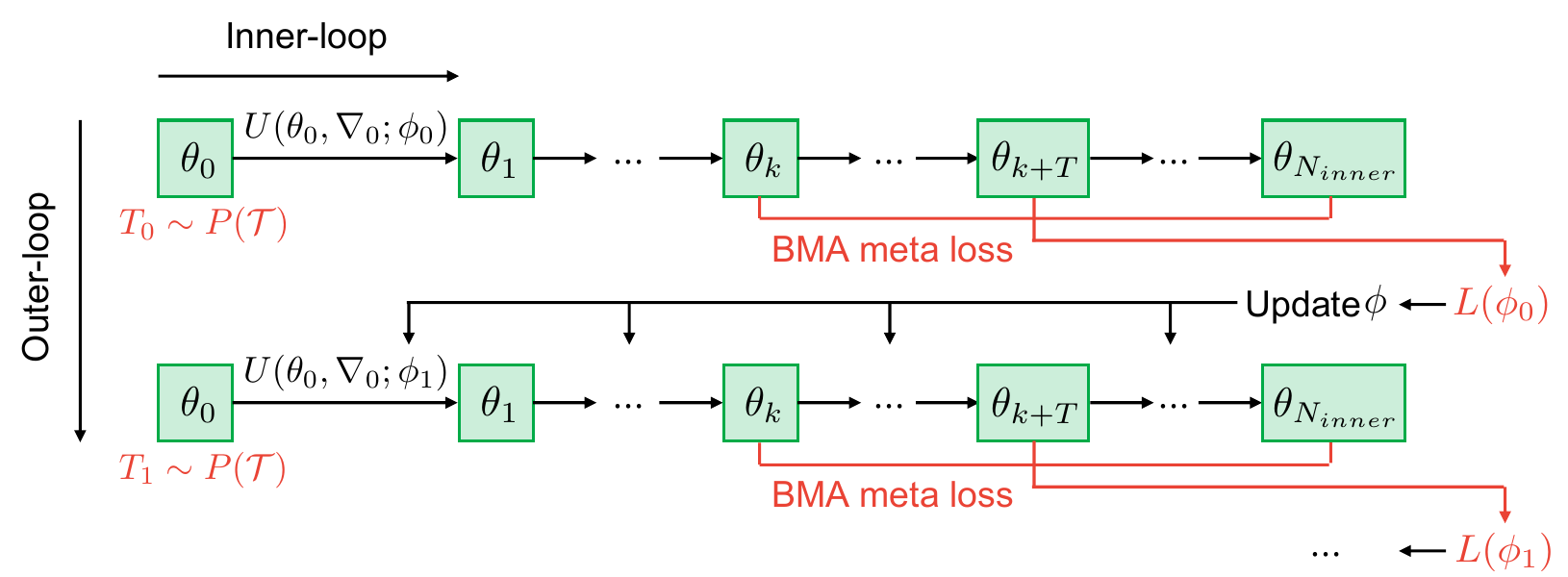}
    \caption{Meta learning procedure of \gls{l2e}.}
    \label{fig:app:concept}
\end{figure}

\subsection{meta-training procedure}

\paragraph{General hyperparameter} For meta-training, we fix the length of the inner loop for all tasks at 3000 iterations. We determine this by monitoring the meta-objective during training and set sufficient length of inner loop for \gls{l2e} to enter to high-density region regardless of tasks. This setting can vary when task distribution is changed. To compute the meta-objective, we collect 10 inner-parameters with a thinning interval of 50 during the last 500 iterations of the inner-loop. We find out that this thinning interval and the number of collected inner parameters do not have a significant impact on the model performance. Also, we use 1000 iterations of outer-loop for training meta-parameter since meta-loss converges after 1000 outer iterations.

\paragraph{Outer optimization} For training meta-parameter, we use Adam 
with learning rate 0.01 and $\beta_1 = 0.9 , \beta_2 = 0.99$. We apply gradient clipping to the gradient of meta-objective to prevent unstable training due to the different gradient scale among tasks.

\subsection{Meta-\gls{sgmcmc} meta-training procedure}
\label{app:sec:metasgmcmctraining}
\paragraph{General setup} For meta-training, we use the same task distribution as \gls{l2e} do to fairly compare generalizability of two methods. We use 1-layer MLP with 10 hidden units for $Q_f$ and $D_f$. Since output of $D_f$ restricted to positive value, we take absolute value of output of $D_f$. We modify some original input preprocessing method different from \citet{gong2018meta}. Firstly, they heuristically find some good values to scale the input for $Q_f$ and $D_f$. This does not work in multi-task training as hyperparameter tuning is task specific, we scale the norm of input of neural networks as \gls{l2e} do. 

For meta-training, we use \gls{tbptt} as the gradient estimator for meta-parameter. We use total 5000 inner-steps as a single problem and use 50 steps for truncation length. After running 5000 steps for one task, we freshly sample the task for training meta-sgmcmc, which is completely different from \citet{gong2018meta} and similar to \gls{l2e}. Since \citet{gong2018meta} used 15 steps for truncation length, we find out that lengthen the truncation length can improve the performance. Also, we tune the band width length and use 100 for stein gradient estimator.

\paragraph{hyperparameters}
For step size, we use 1e-06 since larger step size does not work in our task distribution for Meta-\gls{sgmcmc}. Please refer to \cref{fig:app:meta_loss} for sensitivity of Meta-\gls{sgmcmc} compared to \gls{l2e}. We us 5 for clipping values of $Q_f$ and 100 for $D_f$. For implementing meta-objective, we follow the official implementation of the meta-objective except for number of chains due to memory consumption. We use 2 chains for running Meta-\gls{sgmcmc} on our task-distribution. 

\paragraph{Outer optimization} For training meta-parameter, we use Adam 
with learning rate 0.0003 and $\beta_1 = 0.9 , \beta_2 = 0.99$. We apply gradient clipping to the gradient of meta-objective to prevent unstable training due to the different gradient scale among tasks. We train meta-models for 1000 outer-iteration while saving the meta-models for every 100 outer iterations. Since training Meta-\gls{sgmcmc} is significantly unstable as shown in \cref{fig:app:meta_loss}, we choose the best sampler by evaluating every saved meta-models for fashion-MNIST and CIFAR-10 tasks.

\section{Ablation studies}
\def\textBF#1{\sbox\CBox{#1}\resizebox{\wd\CBox}{\ht\CBox}{\textbf{#1}}}

\begin{table}[t]
    \centering
    \caption{Results of ablation studies. We use 10 samples for ablation studies. \gls{l2e} shows the best performance among other choices. Notable result is that the performance gap between small \gls{l2e} gets larger as the size of task gets bigger. This implies that task distribution of meta-training plays important role for generalization capability. Precond \gls{l2e} shows worse performance in general and diverge in Tiny-ImageNet task. This parameterization shows worse performance and less generalizability than kinetic gradient parameterization.}
    \vskip 0.15in
    \resizebox{0.6\linewidth}{!}{
    \begin{tabular}{lrrrrr}
 \toprule
 Dataset & \gls{l2e} & ACC $\uparrow$ & NLL $\downarrow$ & ECE $\downarrow$  & KLD $\uparrow$ \\

 \midrule
 
\multirow{3}{*}{fashion-MNIST}
& Small \gls{l2e} & 0.9158 & 0.2520 &
0.0120 & 0.0797  \\
& Precond \gls{l2e} & 0.8801 & 0.3461 &
0.0382 & 0.2597  \\
& \gls{l2e} & 0.9166 & 0.2408 &
0.0078 & 0.0507  \\

 \midrule
 
\multirow{3}{*}{CIFAR-10}
& Small \gls{l2e} & 0.8952 & 0.3465 &
0.0610 & 0.5008  \\
& Precond \gls{l2e} & 0.7503 & 0.7950 &
0.1349 & 0.4126  \\
& \gls{l2e} & 0.9009 & 0.3231 &
0.0480 & 0.6263  \\

 \midrule
 
\multirow{3}{*}{CIFAR-100}
& Small \gls{l2e} & 0.6536 & 1.3067 &
0.1237 & 1.4391  \\
& Precond \gls{l2e} & 0.3397 & 2.6761 &
0.1508 & 0.7449  \\
& \gls{l2e} & 0.6702 & 1.2345 & 
0.1224 & 1.2632  \\

 \midrule
 
\multirow{3}{*}{Tiny-ImageNet}
& Small \gls{l2e} & 0.4759 & 2.318 & 0.0672
 & 2.8437   \\
& Precond \gls{l2e}  & -  & -  & -
 & -  \\
& \gls{l2e} & 0.5287 & 1.979 & 0.1478
 & 0.8124   \\

 \bottomrule
    \end{tabular}}     
    \label{app:tab:ablation}
    \vskip -0.1in
\end{table}

In \cref{app:tab:ablation}, we demonstrate the results from our ablations studies. We evaluate how the size of task set can affect the generalization performance of \gls{l2e} and the parameterization choice can make impact on the performance. We use following two variants of \gls{l2e} 
\begin{itemize}
    \item \textbf{Small \gls{l2e}}: meta-trained on only one dataset, using small architectures. In detail, this model only use Fashion-MNIST dataset and channel sizes of 4 and depths of 1 and 2 are possible choices of random configuration.
    \item \textbf{Precond \gls{l2e}}: Parameterization of \gls{l2e} changes from designing kinetic energy gradient to preconditioner. 
\end{itemize}

Firstly, we confirm that \gls{l2e} trained with larger task distribution shows better generalization performance. Although small \gls{l2e} shows decent performance on Fashion-MNIST and CIFAR-10, it shows significant drop of performance in Tiny-ImageNet and CIFAR-100. This implies that further scaling of task distribution can lead \gls{l2e} to show better performance in diverse unseen tasks. Also, we compare the method parameterizing $D(z),Q(z)$ with neural networks as in \citet{gong2018meta}. This shows that our parameterization is significantly better than designing preconditioner. Since preconditioner matrix can only work as multiplier of learning rate or noise scale, it has limitation for its expressivity which shows limitation in \glspl{bnn}.

\section{Experimental Details}
\label{app:sec:experiments}

We use JAX library to conduct our experiments. We use NVIDIA RTX-3090 GPU with 24GB VRAM and NVIDIA RTX A6000 with 48GB for all experiments. We implement meta-training algorithm based on the learned-optimization package~\citep{metz2022practical} with some modifications.

\subsection{Real-world image classification}
\label{app:sec:baseline}
\paragraph{Dataset} We use tensorflow dataset for fashion-MNIST, CIFAR-10 and CIFAR-100. We utilize Tiny-ImageNet with image size of 64x64.

\paragraph{Architecture} We use 1 layer convolution neural network on fashion-MNIST and ResNet56-FRN with Swish activation on Tiny-ImageNet. 
\paragraph{Hyperparameter}
\def\textBF#1{\sbox\CBox{#1}\resizebox{\wd\CBox}{\ht\CBox}{\textbf{#1}}}

\begin{table}[t]
    \centering
    \caption{Hyperparameters for \gls{de}.}\label{app:tab:hyper_de}
    \vskip 0.15in
    \resizebox{0.75\linewidth}{!}{
    \begin{tabular}{lrrrr}
 \toprule
 Method & fashion-MNIST & CIFAR-10 & CIFAR-100 & Tiny-ImageNet  \\

 \midrule
 
Optimizer & SGDM & SGDM & SGDM & SGDM \\
Num models & 100 & 100 & 100 & 10 \\
Total epochs & 10000 & 10000 & 10000 & 2000 \\
Initial learning rate & $0.1$ & $0.1$ & $0.1$ & $0.1$\\
Learning rate schedule & Cosine decay & Cosine decay & Cosine decay & Cosine decay \\
momentum decay & 0.5 & 0.1 & 0.05 & 0.3 \\
Weight decay & $5\times10^{-4}$ & $1\times10^{-3}$ & $5\times10^{-4}$ & $5\times10^{-4}$\\
Batch size & 128 & 128 & 128 & 128 \\

 \bottomrule
    \end{tabular}}
    \vskip -0.1in
\end{table}

\def\textBF#1{\sbox\CBox{#1}\resizebox{\wd\CBox}{\ht\CBox}{\textbf{#1}}}

\begin{table}[t]
    \centering
    \caption{Hyperparameters for \gls{csgmcmc} .}\label{app:tab:hyper_csgmcmc}
    \vskip 0.15in
    \resizebox{0.75\linewidth}{!}{
    \begin{tabular}{lrrrr}
 \toprule
 Method & fashion-MNIST & CIFAR-10 & CIFAR-100 & Tiny-ImageNet  \\

 \midrule

Num burn in epochs & 100 & 100 & 100 & 100 \\
Num models & 100 & 100 & 100 & 10 \\
Total epochs & 5100 & 5100 & 5100 & 1100 \\
Thinning interval & 50 & 50 & 50 & 50\\
exploration ratio & 0.8 & 0.94 & 0.94 & 0.8 \\
Step size & $2\times10^{-6}$ & $1\times10^{-6}$  & $1\times10^{-6}$  & $1\times10^{-6}$ \\
Step size schedule & Cosine & Cosine & Cosine & Cosine \\
Momentum decay & 0.5 & 0.01 & 0.01 & 0.05 \\
Weight decay & $5\times10^{-4}$ & $1\times10^{-3}$ & $1\times10^{-3}$ & $5\times10^{-4}$\\
Batch size & 128 & 128 & 128 & 128 \\

 \bottomrule
    \end{tabular}}
    \vskip -0.1in
\end{table}





\def\textBF#1{\sbox\CBox{#1}\resizebox{\wd\CBox}{\ht\CBox}{\textbf{#1}}}

\begin{table}[t]
    \centering
    \caption{Hyperparameters for \gls{l2e} .}\label{app:tab:hyper_l2e}
    \vskip 0.15in
    \resizebox{0.75\linewidth}{!}{
    \begin{tabular}{lrrrr}
 \toprule
 Method & fashion-MNIST & CIFAR-10 & CIFAR-100 & Tiny-ImageNet  \\

 \midrule

Num burn in epochs & 100 & 100 & 100 & 100 \\
Num models & 100 & 100 & 100 & 10 \\
Total epochs & 5100 & 5100 & 5100 & 1100 \\
Thinning interval & 50 & 50 & 50 & 20\\
Step size schedule & Constant & Constant & Constant & Constant \\
Momentum decay & 0.05 & 0.05 & 0.05 & 0.05 \\
Weight decay & $5\times10^{-4}$ & $5\times10^{-4}$ & $5\times10^{-4}$ & $5\times10^{-4}$\\
Batch size & 128 & 128 & 128 & 128 \\
Step size & $2\times10^{-5}$ & $2\times10^{-5}$ & $1\times10^{-4}$ & $2\times10^{-4}$ \\

 \bottomrule
    \end{tabular}}
    \vskip -0.1in
\end{table}

We report hyperparameters for each methods in \cref{app:tab:hyper_de}, \cref{app:tab:hyper_csgmcmc} and \cref{app:tab:hyper_l2e}. We tune the hyperparameters of methods using \gls{bma} \gls{nll} as criterion with number of 10 samples. For all methods including \gls{l2e}, we tune learning rate(step size), weight decay(prior variance) and momentum decay term. We use zero-mean gaussian as prior distribution for \gls{sgmcmc} methods, so that prior variance is equal to half of the inverse of weight decay. For momentum decay, we grid search over $\alpha \in \{0.01, 0.05, 0.1,0.3,0.5\}$. For weight decay, we also search over $\lambda \in \{1e-04,5e-04,1e-03\}$ to find best configuration. For step size, except for \gls{l2e}, we search over $\epsilon \in \{2e-7, 4e-7, 1e-6, 2e-6, 4e-6, 1e-5\}$. For \gls{l2e}, due to scale of output of meta-learner, we additionaly search over bigger step size including $2e-5$, $1e-4$, $2e-4$.

\paragraph{Baselines}
We use the following as baseline methods
\begin{itemize}
    \item \textbf{Deep ensembles}~\citep{lakshminarayanan2017simple} : This method collects parameters trained from multiple different initialization for ensembling. \gls{de} is often compared with Bayesian methods in recent \glspl{bnn} literature like in \citet{izmailov2021bayesian} in that \gls{de} induce similar function to \gls{hmc} which is golden standard in \glspl{bnn} with \gls{bma}.
    \item \textbf{Cyclical Stochastic Gradient MCMC}~\citep{zhang2020csgmcmc}: \citet{zhang2020csgmcmc} introduced cyclic learning rate schedule to \gls{sgmcmc} for improving exploration of sampler. \gls{csgmcmc} usually shows descent predictive performance comparing to other existing \gls{sgmcmc} methods in large-scale experiments.
    \item \textbf{Meta-\gls{sgmcmc}}~\citep{gong2018meta}: \citet{gong2018meta} proposed to meta-learn curl and diffusion matrix of \gls{sgmcmc} to build sampler that can fastly converge to target distribution with small bias. 
\end{itemize}

\paragraph{Metrics}
Let $p(y|x,\theta) \in [0, 1]^K$ be a predicted probabilities for a given input $x$ with label $y$ and $\theta$ is model parameter. $p^{(k)}$ denotes the $k$th element of the probability vector.
We have the following common metrics on the dataset $\calD$ consists of inputs $x$ and labels $y$:
\begin{itemize}
\item Accuracy (ACC):
\[
\operatorname{ACC}(\calD) = \bbE_{(x,y)\in\calD} \left[ I \left[
    y = \argmax_k p^{(k)}(x)
\right] \right].
\]
\item Negative log-likelihood (NLL):
\[
\operatorname{NLL}(\calD) = \bbE_{(x,y)\in\calD} \left[
    -\log{p^{(y)}(x)}
\right].
\]
\item Expected calibration error (ECE):
Actual implementation of \gls{ece} includes dividing predicted probabilities with their confidence. We use following implementation 
\[
\operatorname{ECE}(\calD, N_{\text{bin}}) = \sum_{b=1}^{N_{\text{bin}}} \frac{
    n_b |\delta_b|
}{
    n_1 + \cdots + n_{N_{\text{bin}}}
},
\]
where $N_{\text{bin}}$ is the number of bins, $n_b$ is the number of examples in the $b$th bin, and $\delta_b$ is the calibration error of the $b$th bin. We use $N_{\text{bin}}=15$ for computing \gls{ece}.

\item Pairwise Kullback-Leibler Divergence(KLD):
Given set of ensemble members $\Theta = \{\theta_i \dots \theta_M\}$, we construct matrix $A$ using pair of ensemble members which have $A_{ij} = p(y|x,\theta_i)\log\frac{p(y|x,\theta_i)}{p(y|x,\theta_j)} $ for a given input $x$ and $y$. We calculate the statistic as follows

\[
\KL(\calD, M) = \bbE_{(x,y)\in\calD} \left[
    { \frac{1}{M(M-1)}\sum_{i \neq j }A_{ij}}
\right].
\]

\item Agreement: The Agreement is defined as the alignment between the top-1 predictions of the HMC and our own predictions. This metric is computed through the following formula:

\[
\operatorname{Agreement}(\calD) = \bbE_{(x,y)\in\calD} \left[ I \left[
    \argmax_k \hat{p}^{(k)}(x) = \argmax_k p^{(k)}(x)
\right] \right],
\]

where $I[\cdot]$ is indicator function and $\hat{p}$ is predictive distribution of HMC. It indicates how well a method is able to capture the top-1 predictions of HMC.

\item Total Variation(Total Var): Total Variation quantifies the total variation distance between the predictive distribution of the HMC and our own predictions averaged over the data points. Specifically, it compares the predictive probabilities for each of the classes as follows

\[
\operatorname{Total Var}(\calD) = \bbE_{(x,y)\in\calD} \left[ \frac{1}{2} \sum_{k} \left|
    \hat{p}^{(k)}(x) - p^{(k)}(x)
\right| \right].
\]

\end{itemize}

\subsection{Computational cost}
\def\textBF#1{\sbox\CBox{#1}\resizebox{\wd\CBox}{\ht\CBox}{\textbf{#1}}}

\begin{table}[t]
    \centering
    \caption{Wall clock time(sec) of \gls{sgmcmc} methods per single epoch. Measured using NVIDIA RTX A6000.}\label{app:tab:wallclock}
    \vskip 0.15in
    \resizebox{0.7\linewidth}{!}{
    \begin{tabular}{lrrrr}
 \toprule
 Method & fashion-MNIST & CIFAR-10 & CIFAR-100 & Tiny-ImageNet  \\
 \midrule
\gls{csgmcmc} & 0.20 & 1.79 & 1.81 & 41.12 \\
Meta-\gls{sgmcmc}& 0.32 & 2.44 & 2.45 & 47.22 \\
\gls{l2e} & 0.32 & 2.64 & 2.66 & 48.43\\

 \bottomrule
    \end{tabular}}
    \vskip -0.1in
\end{table}
 
\cref{app:tab:wallclock} shows the actual computational time for each methods. \gls{l2e} does not incur significant computational overhead.

\begin{figure}[ht]
    \centering
    \includegraphics[width=0.65\textwidth]{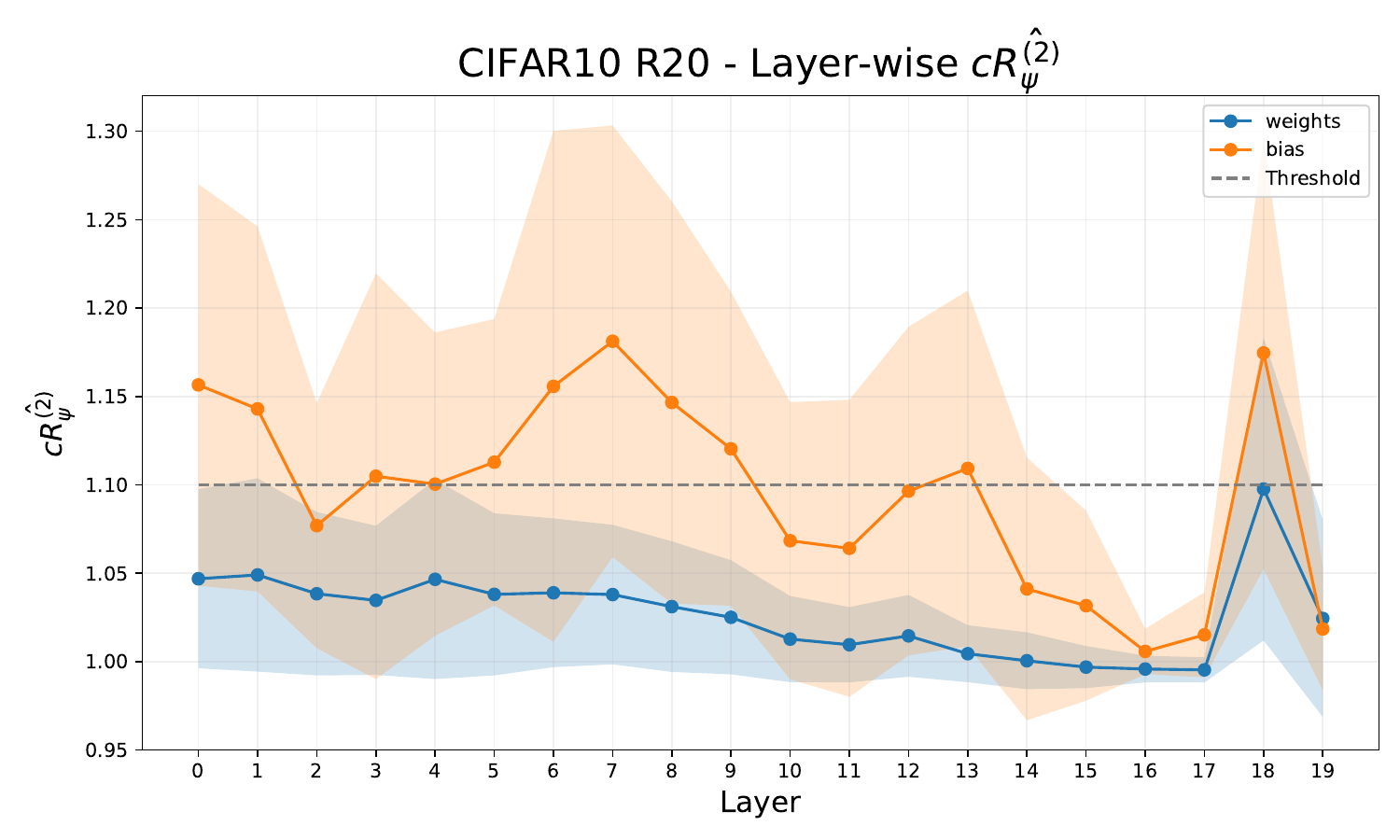}
    \caption{Layerwise convergence of weights and biases on CIFAR-10 ResNet experiment.}
    \label{fig:app:lwiseconv}
\end{figure}

\section{Implementation of Convergence Diagnostics}
\label{app:sec:converge}
For \gls{ess}, we use Tensorflow Probability~\citep{lao2020tfp} library for implementation. We use default parameter of the implementation. \gls{ess} is then normalized to the time consumed for running one thinning interval, which is equivalent to 50 epochs. Since scale of \gls{ess} is too large to report since dimension of neural network parameters are huge for our experiments, we divide it by $10^{-5}$ and report for convenience. 

$\hat{R}$ \citep{gelman1992inference} has been widely employed to measure the convergence of MCMC chain. However, in modern \glspl{bnn}, due to the multimodality in \glspl{bnn} posterior distribution, traditional $\hat{R}$ without further modification is meaningless in the parameter space\citep{sommer2024connecting}. \citet{sommer2024connecting} demonstrated that parameter space convergence should be measured both chain- and layer-wise to fix these issues. Basically, $c\hat{R_{\psi}^{(\kappa)}}$ diagnostic splits a single chain's path into $\kappa$ sub chains for $\hat{R}$ calculation.  Therefore, we report the proportion of parameters with $c\hat{R_{\psi}^{(\kappa)}} < 1.1$ using code implementation of \citet{sommer2024connecting} for our main results. We set $\kappa=2$ considering the number of collected parameters in our experiments. Please refer to \citet{sommer2024connecting} for detailed description of this metric. In \cref{fig:app:lwiseconv}, we also demonstrate layer-wise visualization of $c\hat{R_{\psi}^{(\kappa)}}$. For weights, \gls{l2e} shows good convergence performance in terms of layer-wise $c\hat{R_{\psi}^{(\kappa)}}$ while convergence of biases is mixed across the layers. Considering that weights account for the majority of the parameters, we confirm that \gls{l2e} shows strong convergence in terms of layer-wise $c\hat{R_{\psi}^{(\kappa)}}$.

\section{Limitation of \gls{l2e}}
While effective, \gls{l2e} has some limitations that should be considered for applications. Firstly, \gls{l2e} requires additional meta-training costs. In our experiment, meta-training takes approximately 6 hours on a single NVIDIA RTX A6000 GPU, but it possibly requires more computational cost for meta-training using larger tasks. While L2E demonstrates good performance across various data domains, in \cref{app:tab:ablation}, we observe that the performance of \gls{l2e} is influenced by the size of the meta-training distribution as the scale of the target problem increases. In other words, there is a possibility that we may need to meta-train L2E with larger datasets and architectures to apply L2E to very large models and datasets. 


\end{document}